\documentclass[final]{elsarticle}
\usepackage[a4paper]{geometry}
\usepackage[utf8]{inputenc}
\usepackage{booktabs}

\usepackage{graphicx}
\usepackage{schemata}
\usepackage{color,soul}
\usepackage{amsmath}
\usepackage{multirow}
\usepackage{amsfonts}
\usepackage[table]{xcolor}%
\usepackage{makecell}

\usepackage{soul}
\usepackage{units}
\usepackage{tikz}
\usepackage{standalone, times}
\usetikzlibrary{mindmap,backgrounds}
\usetikzlibrary{positioning}

\definecolor{rootCLR}{RGB}{230,230,230}         %
\definecolor{muslimCLR}{RGB}{241,148,138}       %
\definecolor{gothicCLR}{RGB}{130,224,170}       %
\definecolor{renaissanceCLR}{RGB}{245,203,167}  %
\definecolor{baroqueCLR}{RGB}{169,204,227}      %
\definecolor{leafNodesCLR}{RGB}{100,100,130}      %

\usepackage{algorithm}
\usepackage{algorithmic}

\usepackage{enumitem}
\usepackage{float}

\usepackage[toc,page]{appendix}
\usepackage[modulo]{lineno}
\usepackage{graphicx}
\usepackage{varioref}
\usepackage{array}
\usepackage{amsmath}
\usepackage{booktabs}
\usepackage{multirow}
\usepackage{subcaption}
\usepackage{array,arydshln}
\usepackage{amssymb}
\usepackage{hhline}
\usepackage{times,rotating}
\usepackage{comment}
\usepackage{tikz}
 
 \tikzset{variable/.default=}
\usepackage{tabularx,ragged2e}
\newcolumntype{T}[1]{>{\raggedright\arraybackslash}p{#1}}
\newcolumntype{M}[1]{>{\centering\arraybackslash}m{#1}}

\newcolumntype{L}[1]{>{\raggedright\let\newline\\\arraybackslash\hspace{0pt}}m{#1}}
\newcolumntype{C}[1]{>{\centering\let\newline\\\arraybackslash\hspace{0pt}}m{#1}}
\newcolumntype{R}[1]{>{\raggedleft\let\newline\\\arraybackslash\hspace{0pt}}m{#1}}

\usepackage{tikz}
\usepackage[edges]{forest}
\useforestlibrary{linguistics}
\forestapplylibrarydefaults{linguistics}
\usepackage{pifont}

\usepackage{framed} %
\usepackage{multicol} %
\usepackage{nomencl} %

\usepackage{booktabs}       %
\usepackage{amsfonts}       %
\usepackage{nicefrac}       %
\usepackage{microtype}      %
\usepackage{diagbox} %
\usepackage{dsfont}
\usepackage{bm} %
\usepackage{amsmath}
\usepackage{graphicx}
\usepackage{amsfonts}
\usepackage{subcaption}
\usepackage{stmaryrd}
\usepackage{color, colortbl}
\usepackage{breakcites}
\usepackage{pdfpages}
\usepackage{rotating}
\usepackage{dirtree} %

\usepackage[colorinlistoftodos]{todonotes} %
\usepackage[colorlinks=true, allcolors=blue]{hyperref}

\definecolor{orange}{HTML}{FFC17D}
\definecolor{green}{HTML}{A1D68B}
\definecolor{lightgray}{HTML}{E8E8E8}

\usepackage{hyperref} %

\makenomenclature
\renewcommand*\nompreamble{\begin{multicols}{2}}
\renewcommand*\nompostamble{\end{multicols}}

\journal{Information Fusion}
\usetikzlibrary{arrows.meta,shadows.blur}

\begin{document}

\begin{frontmatter}

\title{%
EXplainable Neural-Symbolic Learning (\textit{X-NeSyL}) methodology to fuse deep learning representations with expert knowledge graphs: the MonuMAI cultural heritage use case
}

\author[a,i,f]{Natalia Díaz-Rodríguez\textsuperscript{*}
\corref{cor1}\corref{eq}}
\author[i,f]{Alberto Lamas\textsuperscript{*}}
\author[a]{Jules Sanchez\textsuperscript{*}}
\author[a]{Gianni Franchi}
\author[c]{Ivan Donadello}
\author[i,f]{Siham Tabik}
\author[a]{David Filliat}
\author[h]{Policarpo Cruz}
\author[f,j]{Rosana Montes}
\author[i,f,g]{Francisco Herrera}
\address[a]{U2IS, ENSTA, Institut Polytechnique Paris and Inria Flowers, 91762, Palaiseau, France}
\address[c]{Free University of Bozen-Bolzano, 39100, Italy}
\address[f]{DaSCI Andalusian Institute %
in Data Science and Computational Intelligence, University of Granada, 18071 Granada, Spain}
\address[g]{Faculty of Computing and Information Technology, King Abdulaziz University, Jeddah, 21589, Saudi Arabia}
\address[h]{Department of Art History, University of Granada, 18071 Granada, Spain}
\address[i]{Department of Computer Science and Artificial Intelligence, University of Granada, 18071 Granada, Spain}
\address[j]{Department of Software Engineering,  University of Granada, 18071 Granada, Spain}
\cortext[cor1]{Equal contribution. Corresponding author e-mail: \texttt{natalia.diaz@ensta-paris.fr}.}

\newpage

\begin{abstract} %
The latest %
Deep Learning (DL) models for detection and classification have achieved an unprecedented performance over classical machine learning algorithms. However, DL models are black-box methods hard to debug, interpret, and certify. DL alone cannot provide explanations that can be validated by a non technical audience such as end-users or domain experts. In contrast, symbolic AI systems that convert concepts into rules or symbols --such as knowledge graphs-- are easier to explain. However, they present lower generalisation and scaling capabilities.  A very important challenge is to fuse DL representations with expert knowledge. One way to address this challenge, as well as the performance-explainability trade-off is by leveraging the best of both streams without obviating %
domain expert knowledge. In this paper, we tackle such problem by considering the symbolic knowledge is expressed in form of a domain expert knowledge graph. %
We present the eXplainable Neural-symbolic learning (\textit{X-NeSyL}) methodology, designed to learn both symbolic and deep representations, together with an explainability metric %
to assess the level of alignment of machine and human expert explanations. The ultimate objective is %
to fuse DL representations with expert domain knowledge during the learning process so it serves as a sound basis for explainability. In particular, X-NeSyL methodology involves the concrete use of two notions of explanation, both at inference and training time respectively: 1) \textit{EXPLANet}: Expert-aligned eXplainable Part-based cLAssifier NETwork Architecture, a compositional convolutional neural network that makes use of symbolic representations, and 2) \textit{SHAP-Backprop}, an explainable AI-informed training procedure that corrects and guides the DL process to align with such symbolic representations in form of knowledge graphs. We showcase X-NeSyL methodology using MonuMAI dataset for monument facade image classification, and demonstrate that with our approach,   it is possible to improve explainability at the same time as performance.    %

 \end{abstract}


\end{frontmatter}

\section{Introduction}
Currently, Deep Learning (DL) constitutes the state-of-the art models in many problems \cite{lecun2015deep,hinton2006reducing,xu2015show,vaswani2017attention,karpathy2014large}. These models are opaque, complex and hard to debug, which makes their use unsafe in critical applications such as healthcare and high-risk scenarios. %
Furthermore, DL often requires a large amount of training data with over-simplified annotations that obviate an important part of centuries-long knowledge from domain experts. At the same time, DL generally uses correlation shortcuts to produce their outputs, which makes them finicky and difficult to correct. On the contrary, most classical symbolic AI approaches are interpretable but do not reach neither similar levels of performance nor scalability.  

Among the potential solutions to clarify the decision process of a DL model, the topic of eXplainable AI (XAI) emerges. Given an audience, an XAI system produces details or reasons to make its functioning clear or easy to understand \cite{arrieta2020explainable, guidotti2018survey}. To make black box Deep Learning methods more interpretable, a large amount of works exposed their vulnerabilities and sensitivity and came up with visual interpretation techniques, such as attribution or saliency maps \cite{zeiler2014visualizing,selvaraju2017grad,olah2017feature}. However, the explanations provided by these methods, often in form of heatmaps, are not always enough, i.e. they are not easy to quantify, correct, nor convey to non technical audiences \cite{Jain2019AttentionIN,wiegreffe-pinter-2019-attention,viviano2021saliency,maguolo2020critic,he2020sample,adebayo2018sanity,kindermans2019reliability}.  

Having both specific and broad audiences of AI models contributes towards inclusiveness and accessibility, both part of the principles for responsible \cite{arrieta2020explainable} and human-centric AI \cite{pisoni2021human}. Furthermore, as advocated in \cite{diaz2020accessible}, broadening the inclusion of different minorities and audiences can facilitate the evaluation of AI models when the objective is deploying human-centred AI systems. 

A very critical challenge is thus to blend DL representations with domain expert knowledge. This leads us to draw inspiration from Neural-Symbolic (NeSy) learning \cite{dAvilaGarcez19NeSy,besold2017neural}, a learning paradigm composed by both neural (or sub-symbolic) and symbolic AI components. An interesting challenge consists of bringing explainability in this fusion through the alignment of such learned and symbolic representations \cite{Bennetot19}.
In order to pursue this idea further, we approach this quest by considering the expert knowledge to be in form of a KG. 

Since our ultimate objective is fusing DL representations and domain expert representations, to fill this gap we propose the eXplainable Neural-symbolic (\textit{X-NeSyL}) learning methodology, to bring explainability in the process. %
X-NeSyL methodology is aimed to make neural-symbolic models explainable, while providing more universal explanations for both end-users and domain experts. X-NeSyL methodology is designed to enhance both performance and explainability of DL, in particular, a convolutional neural network (CNN) classification model. X-NeSyL methodology is constituted by three main components:

\begin{enumerate}
    \item A symbolic processing component to process symbolic representations, in our case we model explicit knowledge from domain experts with knowledge graphs.
    \item A neural processing component to learn neural representations, \textit{EXPLANet}: eXplainable Part-based cLAssifying NETwork architecture. EXPLANet is a compositional deep architecture that allows to classify an object by its detected parts. 
    \item An XAI-informed training procedure, able to guide the model to align its outputs with the symbolic explanation and penalize it accordingly when this is not the case. We propose SHAP-backprop to align the representations of a deep CNN with the symbolic one from a knowledge graph, thanks to a SHAP Attribution Graph (SAG) and a misattribution function. %
   
\end{enumerate}

The election of these components is designed to enhance a DL model by endowing its output with explanations at two levels:
\begin{itemize}
    \item \textit{Enhancement of the explanation at \textit{inference} time}: %
    We extend the classifier inference procedure to not only classify, but also detect what will serve as basis for the explanation. 
    These components should be possible to be specified through the symbolic component, e.g., a knowledge graph that acts as gold standard explanation from the expert. %
    EXPLANet is proposed here to classify an object based on the detected object-parts, and thus, has the role of facilitating the mapping of neural representations to symbols. %
   
    \item \textit{Enhancement of the explanation at \textit{training} time}: %
    We penalize the original model at this second training phase, aimed towards improving the original classifier, thanks to an XAI technique called Shapley analysis \cite{lundberg2017unified} that assesses the contribution of each feature to a model output. %
    \textit{SHAP-backprop} training procedure is presented to adjust the model using a misattribution function that quantifies the error coming from the contribution of features (object-parts) attributed to the %
    output (expressed in a SHAP Attribution Graph, \textit{SAG}) %
    not in agreement with the theoretical contribution expressed by the expert knowledge graph. 
\end{itemize}

Together with the X-NeSyL methodology, this paper contributes an explainability metric to evaluate the interpretability of the model, SHAP GED (SHAP Graph Edit Distance), that measures the degree of alignment between the symbolic (expert) %
and neural (machine) representations. The objective of this metric is to gauge the alignment between the explanation from the model and the explanation from the human target audience that validates it. 

We illustrate the use of X-NeSyL methodology through a guiding use case on monument architectural style classification and its dataset named MonuMAI \cite{lamas2020monumai}. We selected this dataset because it includes object-part-based annotations which make it suitable for assessing  our proposal.

\begin{figure}[htbp!]
    \centering
    \includegraphics[width=\textwidth]{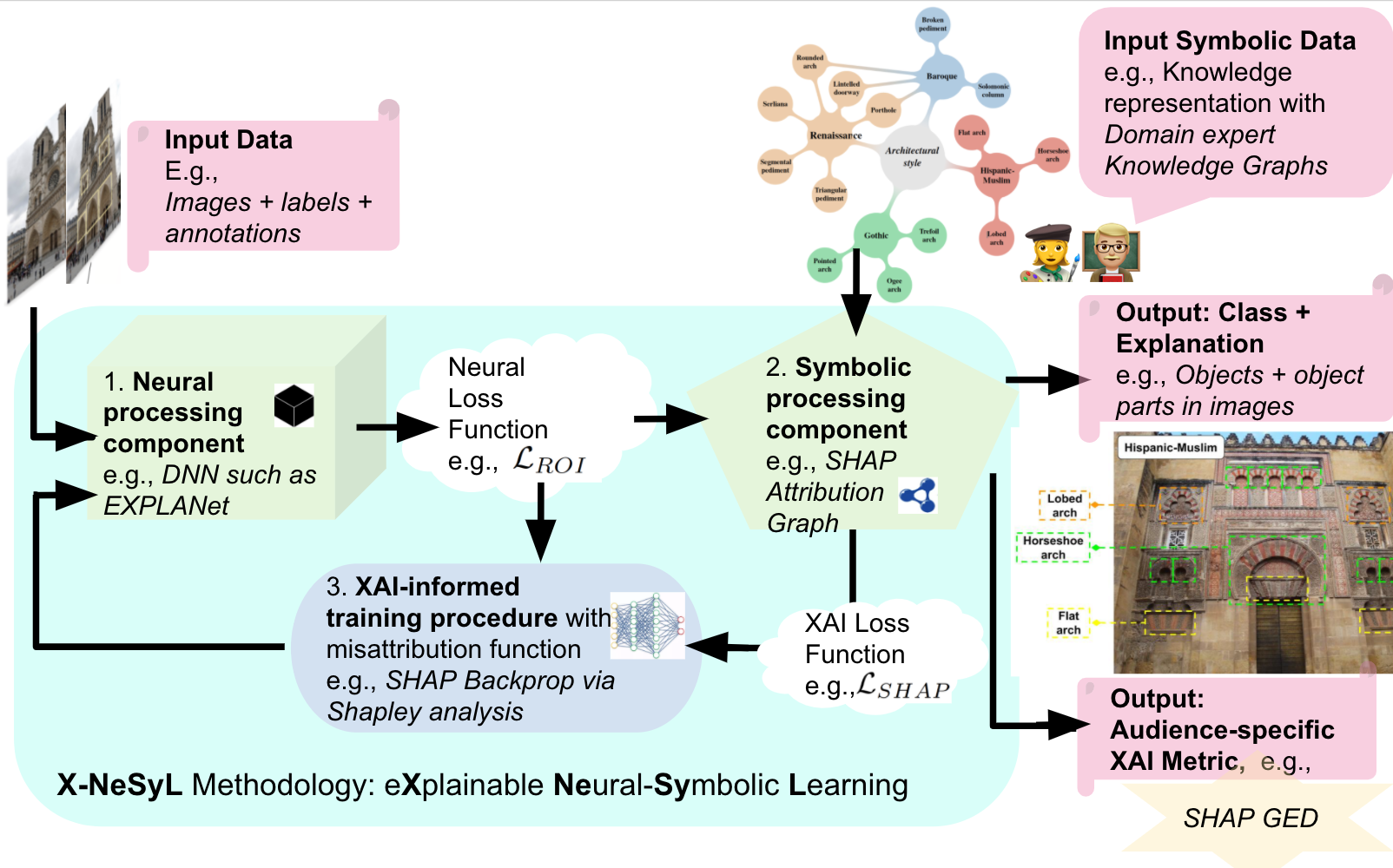}
    \caption{Proposed X-NeSyL methodology for eXplainable Neural-Symbolic learning. We illustrate the components of the methodology with examples of MonuMAI use case processed with \textit{EXPLANet} part-based object detection and classification model, knowledge graphs, and \textit{SHAP-Backprop} training procedure.}
    \label{fig:methodology} %
\end{figure}

The pipeline components of the X-NeSyL methodology are summarized in Fig. \ref{fig:methodology}. They are meant to complete a versatile template architecture with pluggable modular components to make possible the fusion of representations of different nature. X-NeSyL methodology can be adapted to the needs of the use case, and allows the model to train in a continual learning \cite{lesort2020continual} setting. 

The experiments to validate the X-NeSyL methodology make evident the well known interpretability-performance trade-off %
with respect to traditional training with an improvement of %
3.6 \% with respect to the state of the art (\textit{MonuNet} \cite{lamas2020monumai}) on MonuMAI dataset. In terms of explainability, our contributed interpretability metric, SHAP GED, reports a gain of up to 0.38 --from 0.93 to 0.55--. %
The experimental study shows that X-NeSyL methodology %
makes it possible for CNNs to gain explainability and performance. 

The rest of this paper is organized as follows: First we present the literature around XAI, and compositional, part-based classifiers in Section \ref{sec:relatedwork}. We %
present a set of frameworks on Neural-Symbolic integration as a basis and promising body of research to attain XAI %
in Section \ref{sec:nesy}. %
We %
describe %
X-NeSyL methodology in Section \ref{sec:xnesyl}. Its core components are presented therein, Section \ref{sec:representingDomainKnowledgeKG} presents the symbolic component, i.e., how KGs can be used to represent symbolic expert knowledge to be leveraged by a DL model, Section \ref{sec:explanet} presents the neural representation component describing EXPLANet architecture, and Section \ref{sec:SHAP-backprop} the XAI-guided training method \textit{SHAP-Backprop}. X-NeSyL methodology is evaluated through the proposed explainability metric SHAP GED, presented and illustrated in Section \ref{sec:shapged}. The complete methodology pipeline is illustrated through a driving use case on MonuMAI cultural heritage application  %
in %
Section \ref{sec:experiments}. %
Section \ref{sec:futureworks} we discuss %
results, alternative perspectives, and open research avenues for the future. Finally, the Appendix includes additional experiments with an extra dataset, PASCAL-Part. %

\section{Related work: Explainable deep learning and compositional part-based classification }%
\label{sec:relatedwork}

An increasing number of reviews and surveys  is providing diverse classifications of XAI methods \cite{arrieta2020explainable,buhrmester2019analysis,guidotti2018survey}. We particularly focus on attribution methods, i.e. XAI methods that relate a particular output of a DL model to their input variables. The most popular methods are model agnostic and provide visual explanations in form of heatmaps, saliency maps or class activation methods. %

This section reviews three types of XAI attribution  methods \cite{olah2017feature, arrieta2020explainable}, 1) local explanations, 2) saliency maps and 3) compositional part-based classification models.

\subsection{Local explanations}
\label{sec:local_explanations}

These methods are increasingly model agnostic, i.e.,  independent of the underlying black box model. The main intuition behind this type of methods is that they start from any specific point in the input space and explore its neighborhood to understand what has caused the prediction.

One of the  simplest approaches in this context is \textit{LIME} \cite{ribeiro2016should}. It explains the decision of the model by example. That is, in image classification,  LIME  segments  the input image into super-pixels and then generates a data set of perturbed instances by turning some of the super-pixels into gray color. Each perturbed instance is then analyzed by the machine learning model to get a decision. Afterwards, a simple linear locally weighted model is trained on this data set. As result, the super-pixel with the highest weight is presented as explanation. One of the limitations of this approach is that it cannot generalize to unseen instances. 

To increase generalization to unseen images,   the same researchers  introduced \textit{Anchors} \cite{ribeiro2018anchors}. Instead of using a simple linear model, Anchors use reinforcement learning techniques in combination with a graph search algorithm. They express the explanation using simple $if-then$ rules applicable to other instances. 

Following the same philosophy, \textit{Minimal input deformation} \cite{fong2017interpretable}, uses blur as perturbation technique to learn the perturbation mask that minimizes the class score. This can be seen as a counterfactual explanation strategy. The authors showed that the obtained image masks provide better explanations than the obtained by previous gradient-based saliency methods and their variants.

\textit{SHAP} (SHapley Additive exPlanation) \cite{lundberg2017unified} is a framework for interpreting predictions based on coalitional game theory and theoretically optimal Shapley Values. It  explains the prediction of an instance $x$ by computing the contribution of each feature to the prediction. The feature values of $x$ act as players in a coalition. The computed Shapley values tell us how to fairly distribute %
the prediction among the features. A player can be an individual feature value for tabular data, or a set of feature values, i.e., a vector of values. In general, SHAP is not applied directly to images but can be applied to a vector of intermediate features, i.e., the output of a high-level layer of a neural network. %

\textit{DeepLIFT} \cite{shrikumar2017learning} %
main specificity is that it computes importance scores based on differences with a reference (in case of images, a black image). There is a connection between DeepLIFT and Shapley values. The Shapley values measure the average marginal effect of including an input over all possible orderings in which inputs can be included. If we define "including" an input as setting it to its actual value instead of its reference value, DeepLIFT can be thought of as a fast approximation of the Shapley values. It can as well be seen as an extension of SHAP for images. %

\subsection{Explanations using saliency maps}
\label{sec:saliency_explanations}
Saliency maps based post-hoc explanation methods have been a very powerful tool in explaining deep  CNNs, as they propose an easily interpretable map. A saliency map is actually a heatmap that is usually superimposed over the input image to emphasize the pixels of the image that were more important for the prediction. %

\textit{DeconvNet \& Guided Backpropagation} \cite{springenberg2015striving} \cite{zeiler2014visualizing} are the first approaches for saliency maps based methods. DeconvNets \cite{zeiler2011adaptive} consist of running the model backwards to map the activations from intermediate convolution layers back to the input pixel space. %
Guided backpropagation is a variant of the standard deconvolution approach that is meant to work on every type of CNNs, even if no pooling layers are present. The main difference lies on the way ReLU functions are handled in the different cases.

\textit{LRP (layer-wise relevance propagation)}: \cite{bach2015pixel,binder2016layer} introduce a novel way to consider the operation done inside a neural architecture with the concept of \textit{relevance}. Given their definition of relevance and adding certain properties, \textit{relevance} intuitively corresponds to the local contribution to the prediction function f(x). %
The idea of LRP is to compute feature relevance thanks to a backward pass and thus, it yields a pixel-wise heat map.

\textit{PatternNet \& PatternAttribution} \cite{kindermans2018learning} take the previous work a step further by applying a proper statistical framework to the intuition behind. More precisely, they build on slightly more recent work called DTD (deep Taylor decomposition) as introduced in \cite{montavon2017explaining}. The key idea of DTD is to decompose the activation of a neuron in terms of contributions from its inputs. This is achieved using a first-order Taylor expansion around a root point $x_0$. The difficulty in the application of DTD is the choice of the root point $x_0$, for which many options are available. \textit{PatternAttribution} is a DTD extension that learns from data how to set the root point.  This way the function extracts the signal from the data, and it maps \textit{attribution} back to the input space, which is the same idea of \textit{relevance}.
\textit{PatternNet} yields a layer-wise back-projection of the estimated signal to the input space.

\textit{CAM (Class Activation Mapping)} \cite{zhou2016learning} has as goal to leverage the effect of Global Average Pooling %
layers to a localization of deep representations for CNNs. A class activation map for a particular category indicates the discriminative image regions used by the CNN to identify that category. The process for the basic CAM approach is to put a Global Average Pooling layer on top of the convolution network and then perform classification. The layer right before the Global Average Pooling is then visualized. %

\textit{Grad-CAM \& Grad-CAM++} \cite{selvaraju2017grad,chattopadhay2018grad} emerged from the need to go faster than CAM and avoid a training procedure to happen. The idea of class activation mapping is kept, but to build weights on the features maps (convolution layers) it is using a backpropagated gradient from the score given to a specific class. The advantage is that no architectural changes or re-training is needed, contrary to the architectural constrains of the CAM approach.
With a simple change, Grad-CAM can similarly provide counterfactual activations for a specific class. 
Grad-CAM++ expands Grad-CAM with an improved way to process the weights of the feature maps. Grad-CAM is widely used for interpreting CNNs.

\textit{Score-CAM} \cite{wang2020score} is a novel approach in between CAM-like approaches and local explanations. The main idea behind this new approach is that it does not need to backpropagate any signal inside the architecture and as such, only a forward pass is needed. %

\textit{Integrated gradient} \cite{sundararajan2017axiomatic} proposes a new way to look at the issue. The underlying idea relies on the complexity of ensuring that a visualisation is correct besides being visually appealing and making sense. Here it introduces two axioms that attribution methods should follow, called \textit{sensitivity} and \textit{implementation invariance}.

\textit{Sensitivity}: An attribution method satisfies Sensitivity if for every input and baseline that differ in one feature but have different predictions, the differing feature is given a non-zero attribution. 

\textit{Implementation invariance}: Two networks are functionally equivalent if their outputs are equal for all inputs, despite having very different implementations. Attribution methods should satisfy Implementation Invariance, i.e., attributions should always be identical for two functionally equivalent networks.

Some easily applicable sanity checks can be done to verify that a method is dependant on the parameters and the training set \cite{adebayo2018sanity}.
For instance, \textit{integrated gradient} stems from path methods, and verify both precedent axioms. We consider the straight line path (in $R^n$) from %
the base line between $x_0$ and $x$ in the input, and compute the gradients at all points along the path. Integrated gradients are obtained by accumulating these gradients. Specifically, integrated gradients are defined as the path integral of the gradients along the straight line path from the baseline $x_0$ to the input $x$. %

All these methods seem visually appealing, but most of them rely on heuristics about what we want to look at, more or less well defined. Some works have been proposed to check the validity of such methods as tools to understand the underlying process inside neural architectures. %

It has also been highlighted by the research community that saliency methods must be used with caution, and not blindly trusted, given their sensibility to data and training procedures \cite{viviano2021saliency,kindermans2019reliability}. %

\subsection{Compositional Part-based Classification Models}

Compositionality \cite{andreas2019measuring} in computer vision refers to the capacity to represent complex concepts (from objects, to procedures, to beliefs) by combining simpler parts \cite{fodor2002compositionality,andreas2019measuring}. 
Despite CNNs being not inherently compositional, compositionality is a desirable property for them to be learned \cite{stone2017teaching}.
For instance, hand-written symbols can be learned from only a few examples using a compositional representation of the strokes \cite{lake2015human}. The compositionality of neural networks has also been regarded as key to integrate symbolism and connectionism \cite{hupkes2019compositionality,mao2019neuro}.

Part-based object recognition is an %
example of semantic compositionality and a classic paradigm, where the idea is to gather local level information to make a global classification. In \cite{de1999object}, the authors propose a pipeline that first groups pixels into superpixels, then does segmentation at the superpixel-level, transforming this segmentation into a feature vector and finally classifying the global image thanks to this feature vector. Similar work is proposed by \cite{huber2004parts}, where they extend it to 3D data. Here the idea is to classify part of the image into a predefined class, and then use those intermediate predictions to provide a classification of the whole image. The authors of \cite{bernstein2005part} also define mid level features that capture local structure such as vertical or horizontal edges, Haar filters and so on. However they are closer to dictionary learning than to the work we propose in this paper.

One of the most well known object parts detection model is \cite{felzenszwalb2009object}. %
It provides object detection based on mixtures of multiscale deformable part models, based on data mining of hard negative examples with partially labelled data to train a latent SVM. Evaluation is done in PASCAL object detection challenge (PASCAL VOC benchmark). %

Finally, more recently, semi-supervised processes were developed such as \cite{ge2019weakly}. They are proposing a two step neural architecture for fine-grained image classification aided by local detections. The idea is that positive proposal regions are highlighting varied complementary information, and that all this information should be used. In order to do that, first an unsupervised detection model is made by alternatively applying a CRF and Mask-RCNN (given an initial approximation with CAM). Then having a detection model and thus the positive region proposal they are fed to a Bi-Directional LSTM that will produce a meaningful feature vector accumulating information across all regions and then be able to classify the image. It can be seen as unsupervised part-based classification.

\section{Neural-Symbolic (NeSy) Integration models}
 \label{sec:nesy}
One of the most promising approaches to merge deep representations with symbolic knowledge representation that allows explainability to deep neural networks (such as CNNs) is the Neural-Symbolic (NeSy) integration. NeSy integration aims at joining standard symbolic reasoning with neural networks in order to achieve the best of both fields and soften their limitations. A complete survey of this method is provided in \cite{dAvilaGarcez19NeSy,GarcezLG2009}. Indeed, symbolic reasoning is able to work in presence of few data as it constrains entities through relations. However, it has limited robustness in data errors e and requires formalized background knowledge. On the other hand, neural networks are fast and able to infer knowledge. However, they require a lot of data and have limited reasoning properties. Their integration overcomes these limitations and, as stated by \cite{BianchiPHS19,TownsendCM20}, improves the explainability of the learned models. In the following, we present the main NeSy frameworks.

Many NeSy frameworks treat logical rules as constraints to be embedded in a vector space. In most of the cases these constraints are encoded into the regularization term of the loss function in order to maximize their own satisfiability. Logic Tensor Networks %
\cite{SerafiniG16} and Semantic Based Regularization %
\cite{DiligentiGS16} perform the embedding of First-Order Fuzzy logic constraints. The idea is to jointly maximize both training data and constraints. Both methods are able to learn in presence of constraints and perform logical reasoning over data. In Semantic Based Regularization the representations of logical predicates are learnt by kernel machines, whereas Logic Tensor Networks learn the predicates with tensor networks. Other differences regard the dealing of the existential quantifier: skolemization for Logic Tensor Networks, or conjunction of all possible groundings for Semantic Based Regularization. Logic Tensor Networks have been applied to semantic image interpretation by \cite{Donadello17} and to zero-shot learning by \cite{DonadelloS19}. Semantic Based Regularization has been applied, for example, to the prediction of protein interactions by \cite{SaccaTDP14} and to image classification \cite{DiligentiGS16}. Both Logic Tensor Networks and Semantic Based Regularization show how background knowledge is able to i) improve the results and ii) counterbalance the effect of noisy or scarce training data. Minervini at al proposed a regularization method for the loss function that leverages adversarial examples \cite{Minervini018}: the method firstly generates samples that maximize the unsatisfaction of the constraints, then the neural network is optimized to increase their satisfaction. Krieken et al. \cite{KriekenAH19} proposed another regularization technique applied to Semi-Supervised Learning where the regularization term is calculated from the unlabelled data. In the work of \cite{XuZFLB18}, propositional knowledge is injected into a neural network by maximizing the probability of the knowledge to be true.

Other works use different techniques but keep the idea of defining logical operators in terms of differentiable functions (e.g., \cite{TowellS94,GarcezZ99}). Relational Neural Machines \cite{MarraDGGM20} is a framework developed that integrates neural networks with a First-Order Logic reasoner. In the first stage, a neural network computes the initial predictions for the atomic formulas, whereas, in a second stage, a graphical model represents a probability distribution over the set of atomic formulas. Another strategy is to directly inject background knowledge into the neural network structure as done in \cite{danieleKENN}. Here, the knowledge is injected in the model by adding new layers to the neural network that encode the fuzzy-logic operator in a differentiable way. Then, the background knowledge is enforced both at inference and training time. In addition, weights are assigned to rules as learnable parameters. This allows for dealing with situations where the given knowledge contains errors or it is softly satisfied by the data without a priori knowledge about the degree of satisfaction.

The combination of logic programming with neural networks is another exploited NeSy technique. Neural Theorem Prover \cite{RocktaschelR16} is an extension of the logic programming language Prolog where the crisp atom unification is soften by using a similarity function of the atoms projected in an embedding space. Neural Theorem Prover defines a differentiable version of the backward chaining method (used by Prolog) with the result of learning a latent predicate representation through an optimisation of their distributed representations. \textit{DeepProbLog} \cite{manhaeve2018deepproblog} integrates probabilistic logic programming (\textit{ProbLog} \cite{RaedtKT07}) with (deep) neural networks. In this manner, the explicit expressiveness of logical reasoning is combined with the abilities of deep nets.

Finally, NeSy systems have also shown to be effective for learning logical constraints from KGs \cite{dong2019neural}, for inferring causal graphs from time series \cite{yang2019learn}, for learning to explain logic inductive learning \cite{lowe2020amortized} and for generating symbolic explanations of DL models \cite{tiddi2020knowledge,sarker2020wikipedia,ebrahimi2021towards}.

\section{%
EXplainable Neural-Symbolic (X-NeSyL) learning methodology}%
\label{sec:xnesyl}

One challenge of the latest DL models today is producing not only accurate but also reliable outputs, i.e., outputs whose explanations agree with the ground truth, and even better, agree with a human expert on the subject.
X-NeSyL methodology is aimed at filling this gap, and getting model outputs and experts explanations to coincide. In order to tackle the concrete problem of fusing DL representations with domain expert knowledge in form of knowledge graphs, in this section we present the three main ingredients that compose the X-NeSyL methodology: 1) the symbolic knowledge representation component, 2) the neural representation learning component, and 3) the alignment mechanism for both representations to align, i.e., correct the model during training or penalize it when disagreeing with the expert knowledge.

First, in Section \ref{sec:symbolic-component} we present the symbolic component that serves to endow the model with interpretability --which will be in form of knowledge graphs--, then in Section \ref{sec:explanet} the neural representation learning component --that will serve to reach the best performance-- and finally, in Section \ref{sec:SHAP-backprop} the XAI-guided training procedure that makes both components align with SHAP-Backprop during training of the DL model.

\subsection{Symbolic knowledge representation for including human experts in the loop}%
\label{sec:symbolic-component}

Symbolic AI methods are interpretable and intuitive (e.g. they use rules, language, ontologies, fuzzy logics, etc.). They are normally used for knowledge representation. Since we advocate for leveraging the best of both, symbolic and neural representation learning currents, in order to make the latter more explainable, here we choose a simple form of representing expert knowledge, with knowledge graphs. Right after, in order to demonstrate the practical usage of X-NeSyL methodology, we present the running use case using knowledge graphs that will demonstrate the usage of this methodology thorough the paper.

\subsubsection{Knowledge Graphs} %
\label{sec:representingDomainKnowledgeKG}
Different options exist to leverage a KG as a versatile element to convey explanations \cite{lecue2020role}. We inspire ourselves by NeSy frameworks for XAI using ontologies and KGs \cite{bollacker2019extending,Bennetot19,confalonieri2021using}, %
on explanations of image and tabular data-based models and, more broadly, on the XAI literature \cite{guidotti2018survey}. We focused more precisely on attribution methods that try to measure the importance of the different parts of the input toward the output. We provide a formalization of the domain expert data into a semantic OWL2-based KG that is actually leveraged by the detector and classifier DL model.

In this work we present a new training procedure to enhance interpretability of part-based classifier, given an appropriate KG. It is based on Shapley values (or SHAP) \cite{lundberg2017unified} which outputs feature attribution of the various part elements toward the final classification, which we compare with the KG. We use the SHAP information to weight the loss that we backpropagate \cite{le1989handwritten} at training time. 

Alongside standard images and annotations we have in our various datasets, we also have expert knowledge information. This information is usually encoded in knowledge graphs (KGs), such as the one in Figure \ref{fig:taxonomy}. 

A knowledge graph $\mathcal{G}$ is formalized as a subset of triples from $\mathcal{E} \times \mathcal{R} \times \mathcal{E}$, with $\mathcal{E}$ the set of entities and $\mathcal{R}$ the set of relations. A single triple $(e_i, r, e_j)$ means that entity $e_i$ is related to $e_j$ through relation $r$. In the context of  part-based classification, such graph  encodes the $partOf$ relationship (that is, $\mathcal{R} = \{partOf\}$) between elements (\textit{parts}) and the (\textit{whole}) object they belong to. 

Therefore, since we have $|\mathcal{R}| = 1$, and to relate to various attribution functions presented in Section \ref{sec:local_explanations} and \ref{sec:saliency_explanations}, we can see the KG as the theoretical attribution graph. 

The attribution graph encodes whether an element contributes positively or negatively towards a prediction. This way the $KG$ can be rewritten as $KG = (KG_1,...,KG_n)$ (one entry for each macro label) with $KG_i = (KG_{i,1},KG_{i,2},...,KG_{i,m})$ (one entry for each element that is \textit{part-of} the object), $KG_{i,k} = \{-1, 1\}$. %

If a link between an element and a macro (object-level) label exists in the theoretical KG, then it means such element is typical of that label and should count positively toward this prediction, thus, its entry in the matrix representing the $KG$ is equal to $t$. %
If no such link exists, then it means it is not typical of the macro label and should contribute negatively, thus its entry in the matrix $KG$ is equal to $-t$.%
In our case we choose values of the KG edges to be binary, and since we set $|t|$ = 1, %
$KG_{i,k} = \{-1, 1\}$.
Seeing a KG as a feature attribution graph is not the only way to model a KG; we can also encode KGs as ontologies. It is worth mentioning that ontologies can be seen as a set of triples with the format (subject, predicate, object) or (subject, property, value) where edges can have varying semantic meaning following constraints from Description Logics \cite{Baader07}. 

Modeling the graph as an adjacency matrix is not appropriate since architectural style nodes and architectural elements nodes are playing two very different roles. Instead, we model the graph as a directed graph, with edges from the architectural element toward the architectural styles.

\subsubsection{%
A driving use case on cultural heritage: \textit{MonuMAI} architectural style facade image classification}
\label{sec:monumai}

The latest deep learning models have focused on (whole) object classification. We choose part-based datasets as a straight forward way to leverage extra label information to produce explanations that are compositional and very close to human reasoning, i.e., explaining a concept or object based on its parts.

In this work, we interested ourselves in the MonuMAI (Monument with Mathematics and Artificial Intelligence) \cite{lamas2020monumai} citizen science application and corresponding dataset collected through the application, because it complies with the required compositional labels in a object detection task, based on object parts. At the same time, facade classification by pointing relevant architectonic elements is an interesting use case application of XAI. We use this example thorough the article as a guiding application use case that perfectly serves to demonstrate the usage of our part-based model and pipeline for explainability.

The MonuMAI project has been developed at the University of Granada (Spain) and has involved citizens in creating and increasing the size of the training dataset through a smartphone app\footnote{Mobile App available in the project website: \href{https://monumai.ugr.es/}{monumai.ugr.es}}.

\textbf{The MonuMAI dataset}

\begin{figure}[htbp!]
\centering
\begin{minipage}{.4\textwidth}
    \centering
    \includegraphics[width=\linewidth]{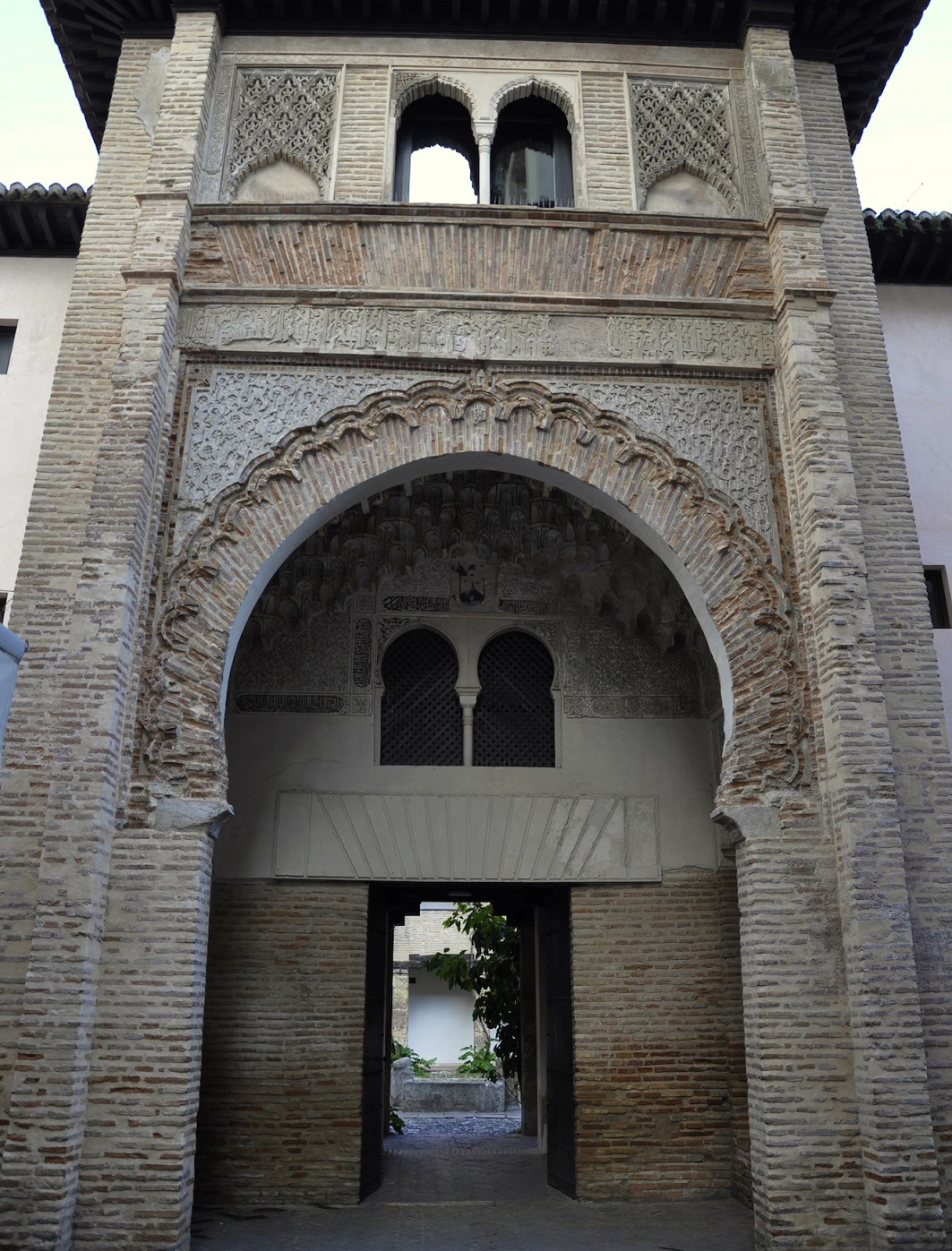}
\end{minipage}%
\begin{minipage}{.4\textwidth}
    \centering
    \includegraphics[width=\linewidth]{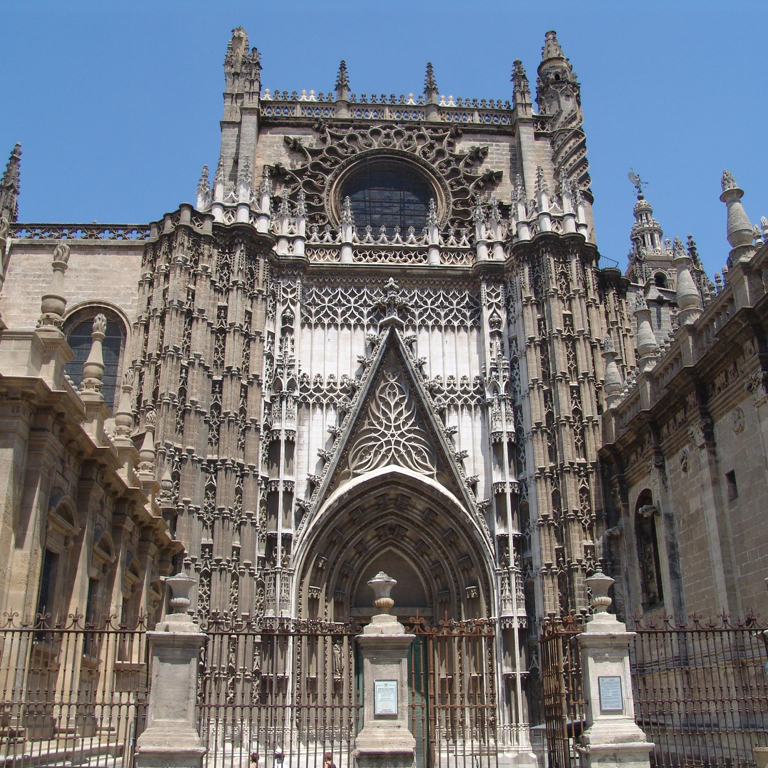}
\end{minipage}
\begin{minipage}{.4\textwidth}
    \centering
    \includegraphics[width=\linewidth]{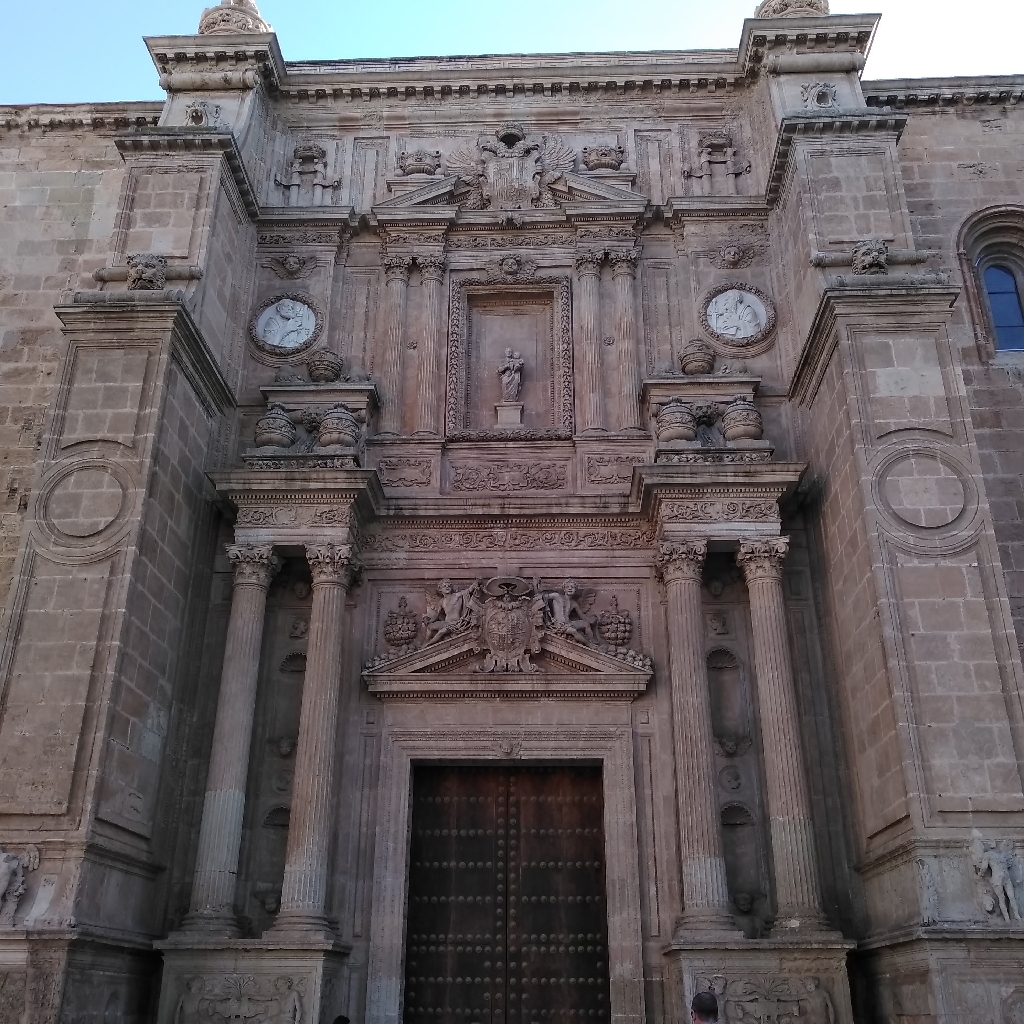}
\end{minipage}%
\begin{minipage}{.4\textwidth}
    \centering
    \includegraphics[width=\linewidth]{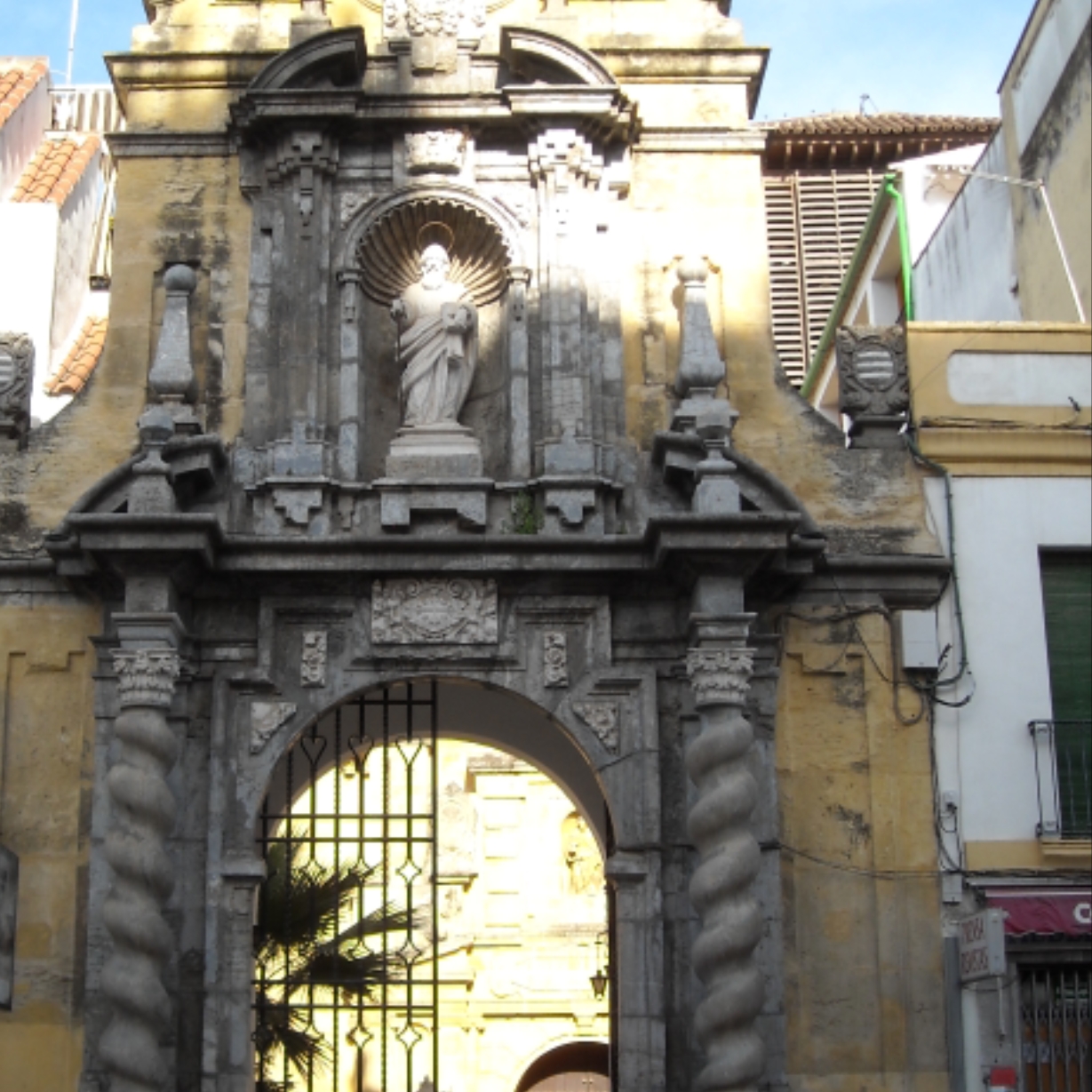}
\end{minipage}
\caption{Extract from MonuMAI dataset. From left to right and top to bottom: hispanic-muslim, gothic, renaissance, baroque (architectural style classes).}
\label{fig:monumai-samples}
\end{figure}

\begin{figure}[htbp!]
\centering
\begin{minipage}{.2\textwidth}
    \centering
    \includegraphics[width=6em, height=6em]{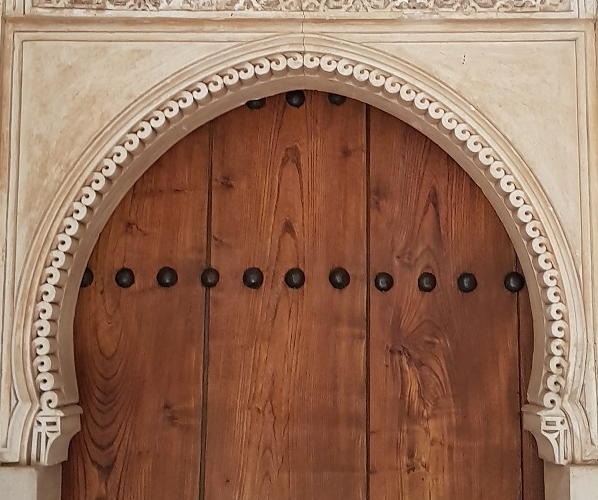}
\end{minipage}%
\begin{minipage}{.2\textwidth}
    \centering
    \includegraphics[width=6em, height=6em]{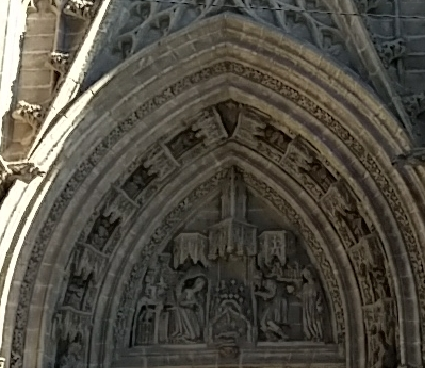}
\end{minipage}
\begin{minipage}{.2\textwidth}
    \centering
    \includegraphics[width=6em, height=6em]{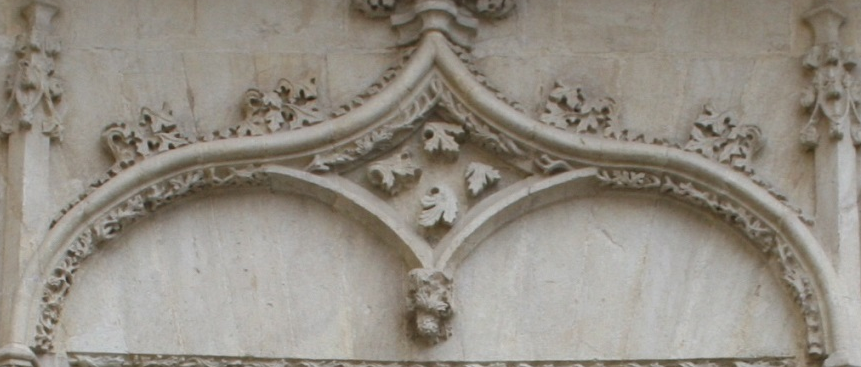}
\end{minipage}%
\begin{minipage}{.2\textwidth}
    \centering
    \includegraphics[width=6em, height=6em]{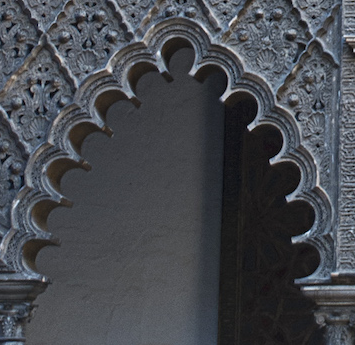}
\end{minipage}
\begin{minipage}{.2\textwidth}
    \centering
    \includegraphics[width=6em, height=6em]{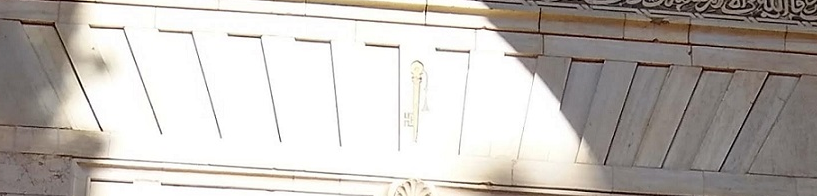}
\end{minipage}
\begin{minipage}{.2\textwidth}
    \centering
    \includegraphics[width=6em, height=6em]{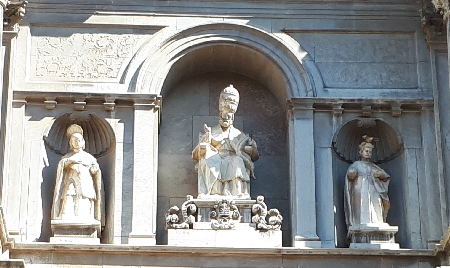}
\end{minipage}
\begin{minipage}{.2\textwidth}
    \centering
    \includegraphics[width=6em, height=6em]{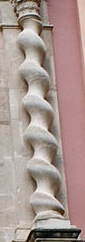}
\end{minipage}
\begin{minipage}{.2\textwidth}
    \centering
    \includegraphics[width=6em, height=6em]{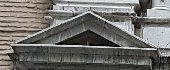}
\end{minipage}
\begin{minipage}{.2\textwidth}
    \centering
    \includegraphics[width=6em, height=6em]{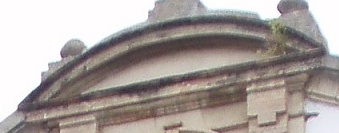}
\end{minipage}
\begin{minipage}{.2\textwidth}
    \centering
    \includegraphics[width=6em, height=6em]{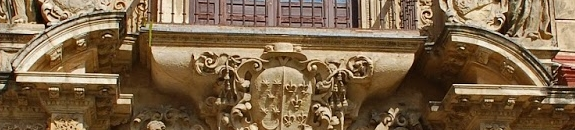}
\end{minipage}
\begin{minipage}{.2\textwidth}
    \centering
    \includegraphics[width=6em, height=6em]{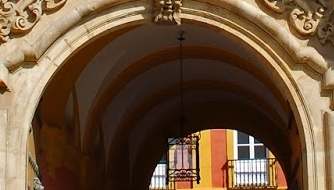}
\end{minipage}
\begin{minipage}{.2\textwidth}
    \centering
    \includegraphics[width=6em, height=6em]{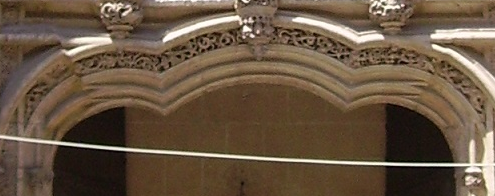}
\end{minipage}
\begin{minipage}{.2\textwidth}
    \centering
    \includegraphics[width=6em, height=6em]{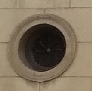}
\end{minipage}
\begin{minipage}{.2\textwidth}
    \centering
    \includegraphics[width=6em, height=6em]{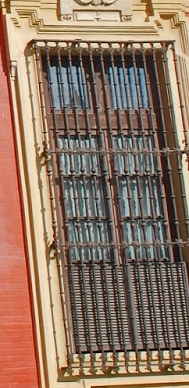}
\end{minipage}
\caption{ 14 architectonic elements (\textit{object parts to be detected}) from MonuMAI dataset used in our experiments as \textit{object-part} explanation to classify a \textit{whole} facade architectural style.  From left to right and top to bottom: horseshoe arch, pointed arch, ogee arch, lobed arch, flat arch, serliana, solomonic column, triangular pediment, segmental pediment, broken pediment, rounded arch, trefoil arch, porthole, lintelled doorway.   The dataset shows that there are architectural elements very distinctive. Even with a very low number of instances in the dataset, object detectors are able to recognize them, e.g., lobed arch, or solomonic column.   }
\label{fig:monumai-samples-part}
\end{figure}

\begin{figure}[htbp!]
    \centering
    \includegraphics[width=16cm]{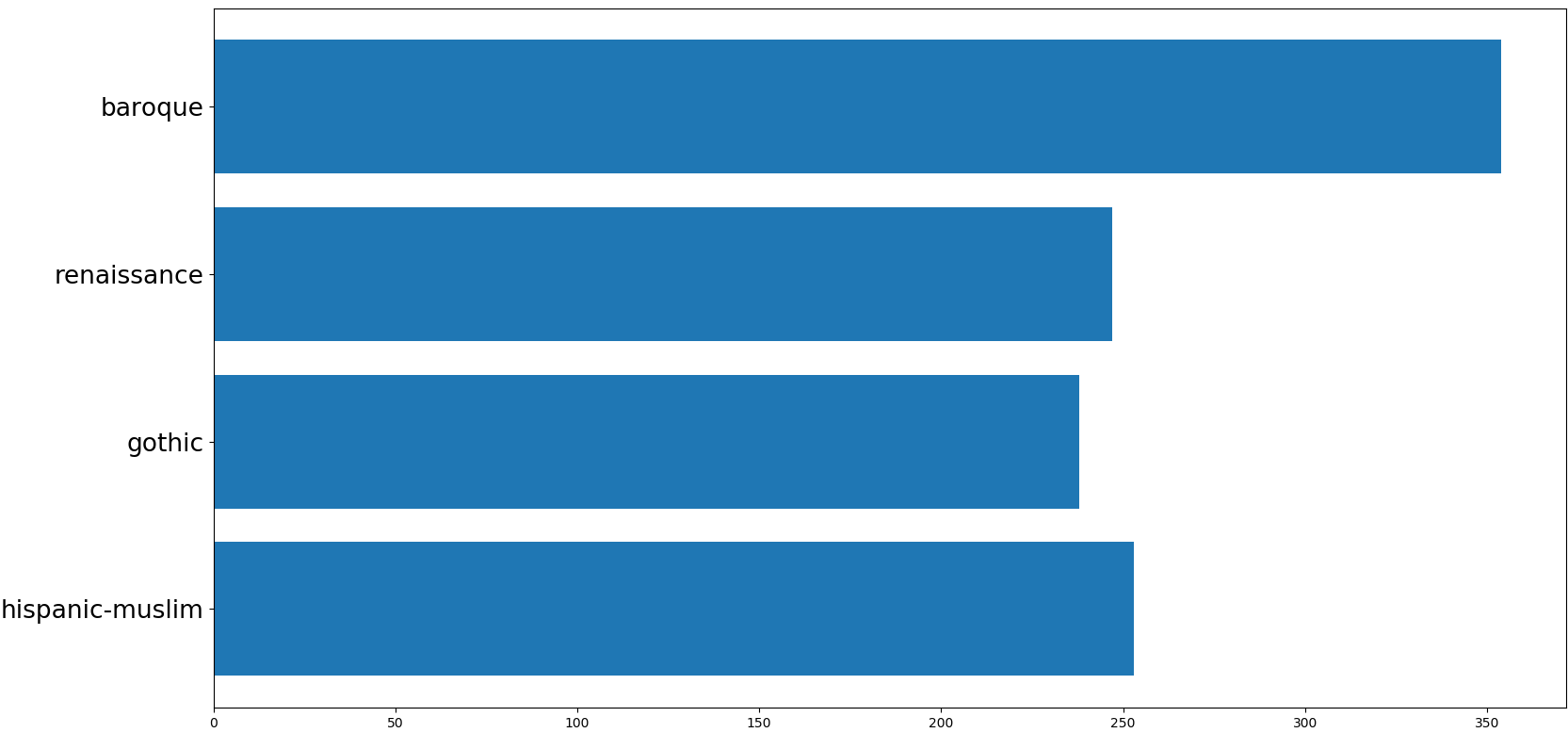}
    \caption{Distribution of style classes in MonuMAI dataset.}
        \label{fig:label_monumai}
\end{figure}

\begin{figure}[htbp!]
    \centering
    \includegraphics[width=17cm]{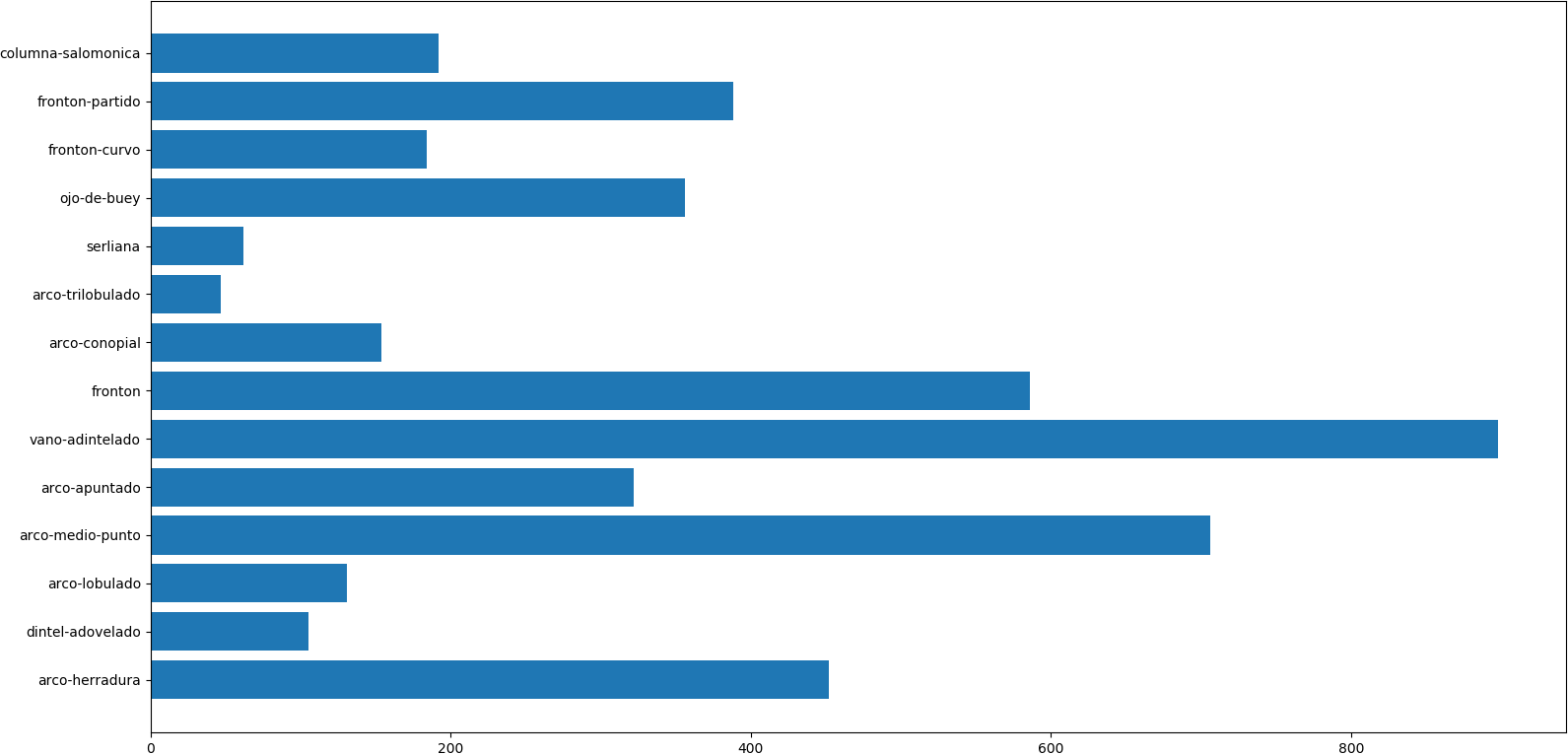}
    \caption{Distribution of object-parts in MonuMAI dataset.}
        \label{fig:part_monumai}
\end{figure}

\begin{figure}[htbp!]
    \centering
    \includegraphics[width=0.7\textwidth]{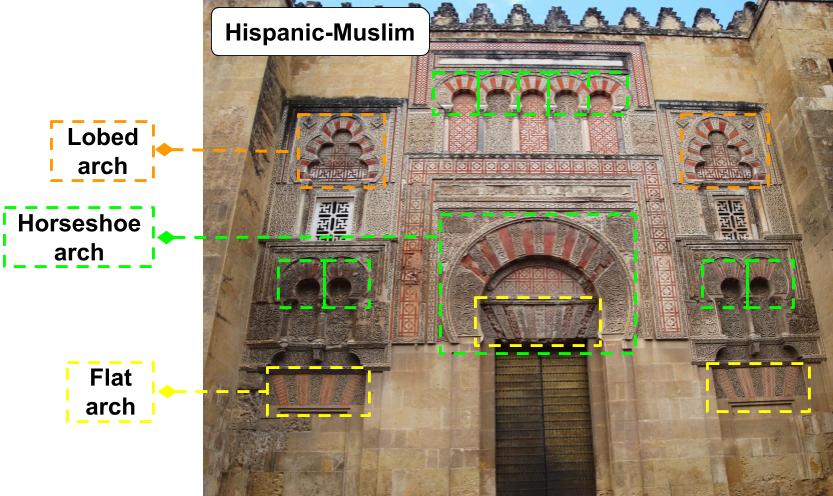}
    \caption{Illustration of the two annotation levels of architectural style and elements on MonuMAI dataset. This image is labeled as Hispanic-Muslim and includes different annotated elements, e.g., two lobed arches (Source: \cite{lamas2020monumai}).}
    \label{fig:labelling}
\end{figure}

\textit{MonuMAI} dataset allows to classify architectural style classification from facade images; it includes $1,092$ high quality photographs, where the monument facade is centered and fills most of the image. Most images were taken by smartphone cameras thanks to the MonuMAI app. The rest of images were selected from the Internet. The dataset was annotated by art experts for two tasks, image classification and object detection as shown Figure \ref{fig:labelling}. All images belong to facades of historical buildings that are labelled as one out of four different styles (detailed in Table \ref{tab:monumai-styles} and Table \ref{tab:monumai-elements}): Renaissance, Gothic, Baroque and Hispanic-Muslim. Besides this label given to an image, every image is labeled with key architectural elements belonging to one of fourteen categories with a total of $4,583$ annotated elements (detailed in Table \ref{tab:monumai-elements}).  Each element is supposed to be typical of one or two styles, and should almost not appear inside facade of the other styles. Examples for each style and each element %
are in Fig. \ref{fig:monumai-samples} and \ref{fig:monumai-samples-part}, while  %
the MonuMAI dataset labels used are shown in Figs. \ref{fig:label_monumai} and \ref{fig:part_monumai}.

\begin{table}[htbp!]
    \centering
    \small
    \begin{tabular*}{0.98\textwidth}{@{\extracolsep{\fill}}lrrr@{}}
    \toprule
    \textbf{Architectural element} & \textbf{Count} & \textbf{Element rate (\%)} & \textbf{Architectural style} \\
    \midrule
    Horseshoe arch          & 452   & 9.86    & Hispanic-muslim \\
    Lobed arch              & 131   & 2.86    & Hispanic-muslim \\
    Flat arch               & 105   & 2.29    & Hispanic-muslim \\
    Pointed arch            & 322   & 7.03    & Gothic \\
    Ogee arch               & 154   & 3.36    & Gothic \\
    Trefoil arch            & 47    & 1.03    & Gothic \\
    Triangular pediment     & 586   & 12.79   & Renaissance \\
    Segmental pediment      & 184   & 4.01    & Renaissance \\
    Serliana                & 62    & 1.35    & Renaissance \\
    Porthole                & 356   & 7.77    & Renaissance/Baroque \\
    Lintelled doorway       & 898   & 19.59   & Renaissance/Baroque \\
    Rounded arch            & 706   & 15.40   & Renaissance/Baroque \\
    Broken pediment         & 388   & 8.47    & Baroque \\
    Solomonic column        & 192   & 4.19    & Baroque \\
    \midrule
    \end{tabular*}
    \caption{Characteristics of the architectural styles dataset, where count is the number of occurrences of an element in the dataset, and element rate is the ratio between the number of occurrences of an element and the total number of all elements.}
    \label{tab:monumai-elements}
\end{table}

\begin{table}[htbp!]
    \centering
    \small
    \begin{tabular*}{0.6\textwidth}{@{\extracolsep{\fill}}lcc@{}}
        \toprule
        \textbf{Architectural style} & \textbf{\#Images} & \textbf{Ratio} (\%) \\
        \midrule
        Hispanic-Muslim         & 253   & 23.17 \\
        Gothic                  & 238   & 21.79 \\
        Renaissance             & 247   & 22.62 \\
        Baroque                 & 354   & 32.42 \\
        \midrule
    \end{tabular*}
    \caption{Characteristics of MonuMAI architectural style classification dataset (\#images represents the number of images).}
    \label{tab:monumai-styles}
\end{table}

Apart from MonuMAI dataset, and in order to draw more general conclusions on our work, we used a dataset with similar hierarchy to MonuMAI. Additional results for PASCAL-Part \cite{chen2014detect} dataset are in the Appendix.

\textbf{MonuMAI's Knowledge Graph}

The original design of MonuMAI dataset and MonuNet baseline architecture \cite{lamas2020monumai} use the KG exclusively as a design tool to visualize the architectural style of a monument facade based on the identified parts, but it is not explicitly used in the model. In contrast, we change that to go further, in order to guarantee a reproducible and explainable decision process that aligns with the expert knowledge. We will see in Section \ref{sec:shapbackprop} how KGs can be used in a detection + classification architecture, during training, since EXPLANet is designed to incorporate the knowledge in the KG. Besides the trust gain, we aim at easing the understanding of flaws and limitations of the model, along with failure cases. This way, requesting new data to experts would be backed up by proper explanations and it would be effortless to target new and relevant data collection.

The KG corresponding to MonuMAI dataset has only fourteen object classes and four architectural styles. %
Each architectural element is linked to at least one style. Each link between the two sets symbolizes that an element is typical and expected in the style it is linked to.

\begin{itemize}
    \item \textit{Renaissance}: rounded arch, triangular pediment, segmental pediment, porthole, lintelled doorway, serliana.
    \item \textit{Baroque}: rounded arch, lintelled doorway, porthole, broken pediment, solomonic column.
    \item \textit{Hispanic-muslim}: flat arch, horseshoe arch, lobed arch.
    \item \textit{Gothic}: trefoil arch, ogree arch, pointed arch.
\end{itemize}

\begin{figure}[!htbp]
    \centering
    \centering
\begin{tikzpicture}
    
    \begin{scope}[mindmap, concept color=rootCLR, text=black, font=\LARGE\bfseries]
        \node [concept, scale=0.85] {\textit{Architectural style}}
            child [grow=-12, level distance=130, concept color=muslimCLR, font=\large\bfseries]{
                node [concept, scale=1.05] (muslim) {Hispanic-Muslim}
                    child [grow=120, level distance=75, font=\normalsize\bfseries]
                        {node [concept, scale=1.05] (flat) {Flat arch}}   %
                    child [grow=20, level distance=90, font=\normalsize\bfseries]{
                        node [concept, scale=1.05] (qin) {Horseshoe arch}   %
                    }
                    child [grow=-90, level distance=95, font=\normalsize\bfseries]{
                        node [concept, scale=1.05] (lob) {Lobed arch}   %
                    }
            }
            child [grow=-115, level distance=130, concept color=gothicCLR, font=\large\bfseries]{
                node [concept, scale=1.05] (gothic) {Gothic}
                    child [grow=0, level distance=80, font=\normalsize\bfseries]
                                {node [concept, scale=1.05] (trefoil) {Trefoil arch}}    %
                    child [grow=-70, level distance=80, font=\normalsize\bfseries]{
                        node [concept, scale=1.05] (qin2) {Ogee arch}   %
                    }
                    child [grow=-150, level distance=90, font=\normalsize\bfseries]{
                        node [concept, scale=1.05] (qin2) {Pointed arch}    %
                    }
            }
            child [grow=-195, level distance=130, concept color=renaissanceCLR, font=\Large\bfseries]{
                node [concept, scale=1.05] (ren) {Renaissance}
                    child [grow=-90, level distance=90, font=\normalsize\bfseries]
                        {node [concept, scale=1.05] (qin) {Triangular pediment}}   %
                    child [grow=-150, level distance=95, font=\normalsize\bfseries]
                        {node [concept, scale=1.05] (qin2) {Segmental pediment}}    %
                    child [grow=150, level distance=95, font=\normalsize\bfseries]
                        {node [concept, scale=1.05] (qin2) {Serliana}}      %
                    child [grow=105, level distance=115, font=\normalsize\bfseries]{
                        node [concept, scale=1.05] (elem1) {Rounded arch}   %
                    }
                    child [grow=70, level distance=75, font=\normalsize\bfseries]{
                        node [concept, scale=1.05] (elem2) {Lintelled doorway}   %
                    }
                    child [grow=-335, level distance=90, font=\normalsize\bfseries]
                        {node [concept, scale=1.05] (elem3) {Porthole}}     %
            }
            child [grow=70, level distance=130, concept color=baroqueCLR, font=\large\bfseries]{
                node [concept, scale=1.05] (baroque) {Baroque}
                    child [grow=100, level distance=80, font=\normalsize\bfseries]{
                        node [concept, scale=1.05] (qin) {Broken pediment}  %
                    }
                    child [grow=-15, level distance=80, font=\normalsize\bfseries]
                        {node [concept, scale=1.05] (qin2) {Solomonic column}}
            };
    \end{scope}
    
    \begin{pgfonlayer}{background}
        \draw (elem1) to[circle connection bar switch color=from (renaissanceCLR) to (baroqueCLR)] (baroque);
        \draw (elem2) to[circle connection bar switch color=from (renaissanceCLR) to (baroqueCLR)] (baroque);
        \draw (elem3) to[circle connection bar switch color=from (renaissanceCLR) to (baroqueCLR)] (baroque);
    \end{pgfonlayer}

\end{tikzpicture}
    \caption{Simplified MonuMAI knowledge graph constructed based on art historians expert knowledge \cite{lamas2020monumai}.}
    \label{fig:taxonomy}
\end{figure}
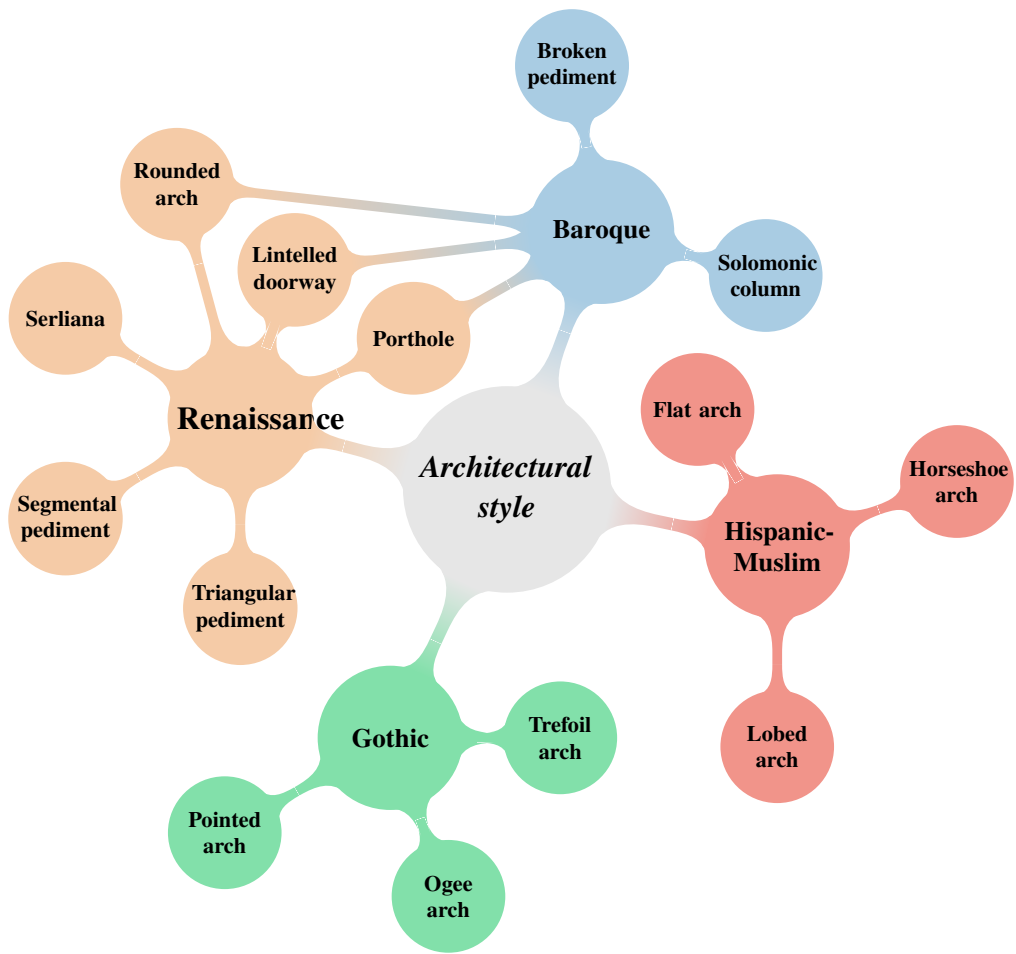

MonuMAI's KG is depicted in Figure \ref{fig:taxonomy}, where the root is the Architectural Style class (which inherits from the \textit{Thing} top-most class in OWL). Note there is one more dimension in the KG, the \textit{leaf} level of the original MonuMAI graph in \cite{lamas2020monumai} that represents some characteristics of the architectural elements, but it is not used in the current work. %

We also explored the possibility of rewriting the looser structure captured in the KG as an ontology, using the OWL2 format. We did not limit ourselves to copying the hierarchy of the original KG, but rather added some categories to keep the ontology flexible to allow further expansions in the future. %
Three main classes are modelled in this ontology: A \textit{Facade} represents an input image as a concept. A facade is linked to one and only one\footnote{In this study, as in MonuMAI, we represent the predominant one. Future work could consider the blend of more than one present style.} \textit{ArchitecturalStyle} through the relation \textit{exhibitsArchStyle}, for which four styles can we used (others could be added by defining new classes). 
A facade can be linked to any number of \textit{ArchitecturalElement} identified on it through the relation (i.e. OWL \textit{object property}) \textit{hasArchElement}. 

\textit{ArchitecturalElement} represents the class of architectural elements identified before, and is divided in subcategories based on the type of elements such as "Arch" or Window". This subcategorization, which does not exist in the original KG, was designed with the possibility of adding constraints between subcategories, such as an "arch" is probably higher in space than a "column", or %
an arch's lowest point is higher than a column's lowest point. Such geometrical or spatial constraints were not explored further, as it required extra expertise modelling from architecture experts, but could be easily added in future work.

Finally, the concept \textit{ArchitecturalElement} is linked to an \textit{ArchitecturalStyle} object through the object property \textit{isTypicalOf}.

This multi-type edge ontology formulation allows us to see the problem of style classification as a problem of KG edge detection  or link prediction among two nodes, in our case, between a facade instance and a style instance. This approach was unsuccessful (discussed in Section \ref{sec:lessonsLearnedAndFutureWork}).

The KG formulation presented in Section \ref{sec:representingDomainKnowledgeKG} can be seen as a semantic restriction of the ontology we propose, where we kept only the triples including \textit{isTypicalOf} relation and expanded the KG with a virtual relation \textit{isNotTypicalOf}, to link together all elements with all the styles.
This way the KG is a directed graph with edges going from the architectural element toward the architectural style. Because we restrict ourselves to only one relational object property and its inverse, the edges bear either \textit{positive} or \textit{negative} information, which motivates our modeling choice of having value $\pm 1$ for formulated $K_{i,j,k}$ edges.

\subsection{%
\textit{EXPLANet}: Expert-aligned eXplainable Part-based cLAssifier NETwork Architecture}
\label{sec:explanet}

 Previous section detailed the symbolic representation mechanism within the X-NeSyL methodology. While KGs serve the purpose of interpretable knowledge, in this section we present the neural representation learning component, mainly responsible for high performance in today's AI systems.

Our ultimate goal in this work is making DL models more trustworthy when it comes to the level of their explanations, and their agreement with domain experts. We will thus follow a human-in-the-loop \cite{holzinger2019interactive} approach. 

Typically, to identify the class of a given object, e.g., an aeroplane, a human first identifies  the key parts of that object, e.g., left wing, right wing, tail; then, based on the combination of these elements and the importance of each single element, he/she concludes the final object class. 

We focus on compositional part-based classification because it provides a common framework to assess part- and whole object based explanations.  To achieve this we want to enforce the model to align with a priori expert knowledge. In particular, we built a new model called \textit{EXPLANet}: Expert-aligned eXplainable Part-based cLAssifier NETwork Architecture, whose design is inspired by the way humans identify the class of an object. 

\begin{figure}[htbp!]
    \centering
    \includegraphics[width=\textwidth]{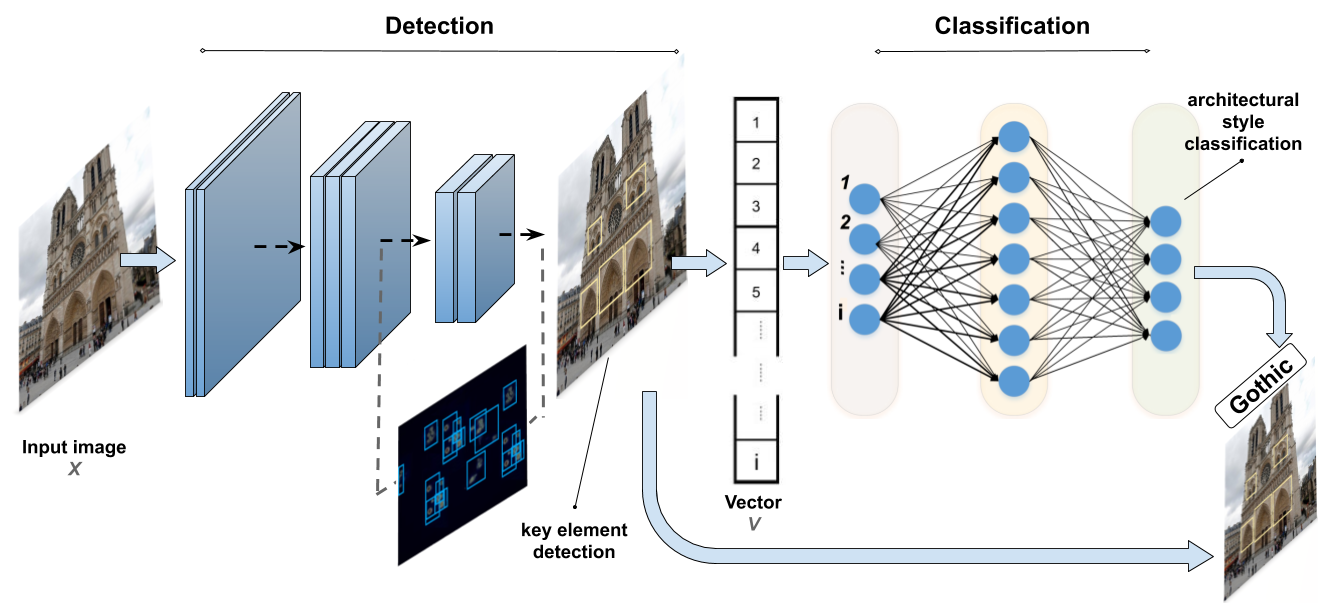}
    \caption{Proposed EXPLANet architecture processing examples from MonuMAI dataset.}
    \label{fig:EXPLANet}
\end{figure}

EXPLANet, is a two-stage classification model as depicted in Figure \ref{fig:EXPLANet}. The first stage detects the  object-parts  present in the input image and outputs an embedding vector that encodes the importance, quantity and combinations of the detected object-parts. This information is used by the second stage to predict the class of the whole object present in the input image. More precisely: 
\begin{enumerate}

 \item %
 The first stage is a detection module, which can be an object detector such as Faster R-CNN \cite{ren2015faster} or RetinaNet \cite{lin2017focal}. Let us consider that there are $n$ object-part classes. This module is trained to detect the key object-part classes existent in the input image, and outputs $M$ predicted regions. Each one is represented by its bounding box coordinates and a vector of size $n$ representing the probability of the $n$ object-part classes.
Let us denote %
$p_m \in \mathbb{R}^n$ (with $m\in [1,M]$) %
the probability vector of detecting object-part $m$.
First we process all $p_m$ by setting non maximal probabilities to zero, and denoting this new score $p'_m$, %
being $p'$ also a vector of size $n$. %
Let us denote vector $v$ the final  descriptor of the image. We build $v$ by accumulating the probabilities of $p'_m$ such that:  
\begin{equation}
    v = \sum_{m=1}^M p'_m \qquad, \quad v \in \mathbb{R}^n
\end{equation}
where vector $v$ aggregates the confidence of each predicted object-part.  $M$ is the amount of bounding box regions predicted. In the case of MonuMAI dataset, to classify 4 object classes (i.e., architectural styles), we have $n=14$ (object-part) labels, but this can be extended to an arbitrary $n$ and $M$.   

Large values in $v$ mean that the input image contains a large number of object-part $i$ elements with a high confidence prediction, whereas a low value means that predictions had low confidence. %
Intermediate values are harder to interpret, as they could be a %
small amount of high confidence predictions or a %
large amount of low confidence predictions, but the idea is that there is probably some %
objects of these kinds in the image. %
This \textit{object-parts vector} can be seen as tabular data %
where each object part can be considered a feature (to be explained later by an XAI method). We will see in next section how a SHAP analysis can study the contribution of each actual object part present in the image to the actual final object classification.

 Note that this probability aggregating scheme is for Faster R-CNN. For each object detected, Faster R-CNN keeps only one prediction (the most likely), whereas RetinaNet keeps all predictions (the whole probability vector).  
We found this to be more stable for training the RetinaNet framework. %

\item The second stage of EXPLANet is a classification network, which is actually a two-layer multi-layer perceptron (MLP), that uses the embedding information (i.e., takes the previous detector output as input) to perform the final classification. This stage outputs the %
final object class based on the importance %
of the present key object parts detected in the input image.
\end{enumerate}

The goal of such design is to facilitate the reproduction of the thought process of an expert, which is to first localize and identify key elements (e.g., in the case of architectural style classification of a facade, various types of arches or columns) and then use this information to deduce the final class (e.g., its overall style). 
However, EXPLANet architecture alone does not control for expert knowledge alignment. Next section introduces the next step of the pipeline, an XAI based training procedure and loss function to actually verify that this happens, and when this is not the case, correct the learning.

\subsection{%
\textit{SHAP-Backprop}: An XAI-informed training procedure and XAI loss based on the SHAP attribution graph (\textit{SAG})}%
\label{sec:SHAP-backprop}

 After having presented the symbolic and neural knowledge processing components of X-NeSyL, we proceed to detail the XAI-informed training procedure to make the most of the best of both worlds, interpretable representations, and deep representations. %
 
More concretely, this section presents how to use a model agnostic XAI technique to make a DL (CNN-based) model more explainable by aligning the test-set feature attribution with the expert theoretical attribution. Both knowledge bases will be encoded in KGs. 

\subsubsection{SHAP values for explainable AI feature contribution analysis}
\label{sec:shap_values}

SHAP is a local explanation method \cite{lundberg2017unified,molnar2020interpretable} that for every singular prediction it assigns to each feature an importance value regarding the prediction. It tells if a feature contributed to the current prediction  and  gives information about how strongly it contributed. These are the Shapley values of a conditional expectation function of the original model. In our case we computed them with Kernel SHAP \cite{lundberg2017unified}.

Similarly to LIME \cite{ribeiro2016should}, Kernel SHAP is a model agnostic algorithm to compute SHAP values.  In LIME, the loss function,  weighting kernel and regularization term are chosen heuristically, while in SHAP they are chosen in a way that they satisfy the SHAP properties. See details in \cite{lundberg2017unified}.

The idea of computing SHAP is to check whether object parts have the expected importance on the object class prediction (e.g. whether the presence of a horseshoe arch contributes to Hispanic-Muslim class). SHAP computation happens always in a per class basis, as the computation is regarding binary classification (belonging to class $C$ vs not).

In our part-based pipeline, we apply SHAP at the tabular data level, i.e., after the aggregation function. As such, SHAP's only input is the feature vector that contains the information about parts detected before. Throughout this section, when we refer to feature value, we refer to this feature vector, and a feature value means one entry of this vector. As such, each feature value encodes the information about one element (either an architectural element for MonuMAI or an object part for PASCAL-Part) from our knowledge model. The final class prediction performed afterwards is done by the classification module of our part-based model, given such a feature vector.

\begin{figure}[htbp!]%
\centering
\begin{minipage}{.75\textwidth}
    \centering
    \includegraphics[width=\linewidth]{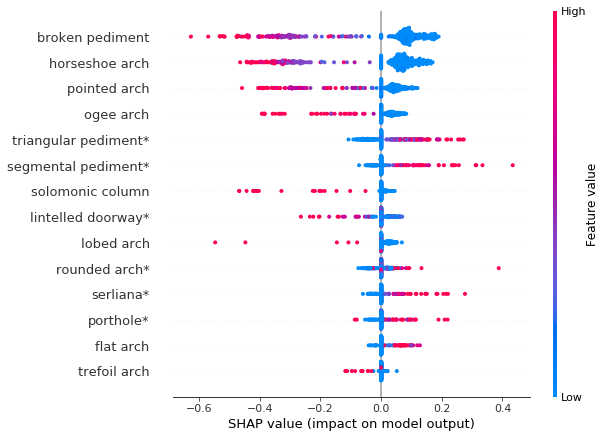}
    \label{fig:shap_R}
\end{minipage}
 \hfill{}%
\begin{minipage}{.75\textwidth}
    \centering
    \includegraphics[width=\linewidth]{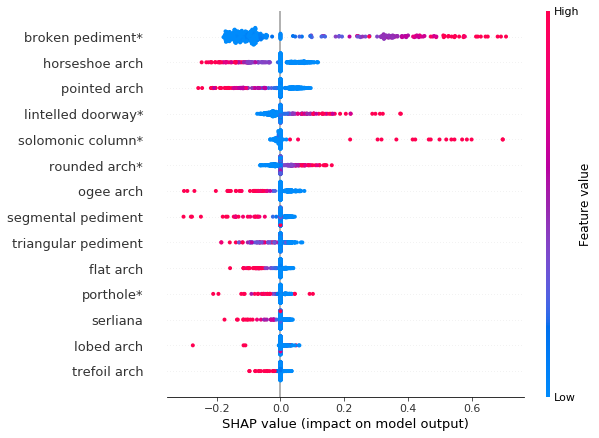}
    \label{fig:shap_B}
\end{minipage}
\caption{SHAP analysis (global explanation as %
SHAP summary plot) for Renaissance (up) and Baroque (down) architectural styles. Each feature (horizontal row) represents a global summary of the distribution of feature importance over the validation set.   Features with asterisk in their name represent the expert ground truth (nodes with a positive value $t$ in the expert KG). Example: In this case, there are features not representing the GT for the class Renaissance, e.g., %
\textit{flat arch} (\textit{dintel adovelado}) is contributing to the prediction, while the KG says it should not. Other than this, generally, datapoints reflecting the GT for an accurate model such as this one, acknowledge properly attributed object part contributions to the classification in the right side of the plot, in pink.  }
\label{fig:shap_RB}
\end{figure}

\begin{figure}[htbp!]%
\centering
\begin{minipage}{.75\textwidth}
    \centering
    \includegraphics[width=\linewidth]{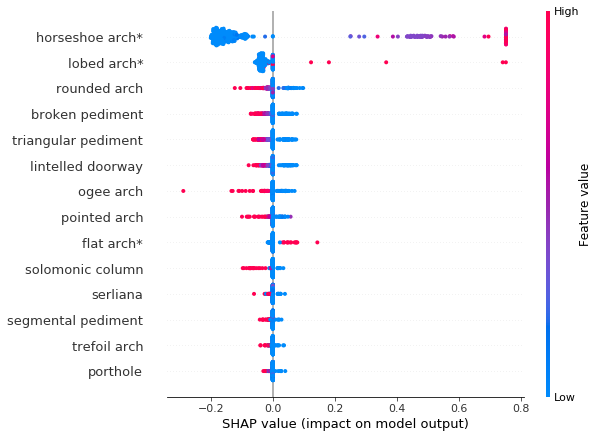} %
    \label{fig:shap_M}
\end{minipage}%
 \hfill{}
\begin{minipage}{.75\textwidth}
    \centering
    \includegraphics[width=\linewidth]{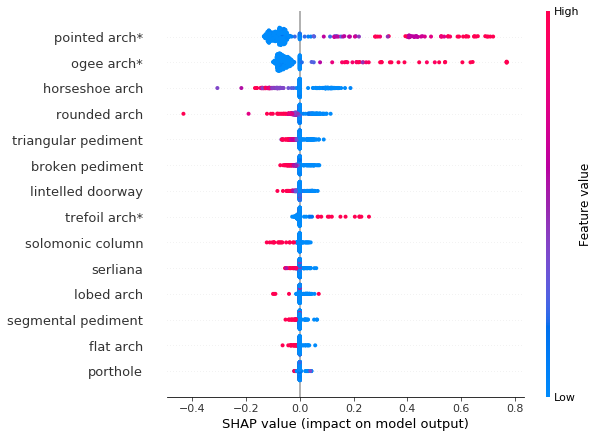}
    \label{fig:shap_G}
\end{minipage}%
\caption{SHAP analysis (global explanation as %
SHAP summary plot) for Muslim (up) and Gothic (down) architectural styles. Each feature (horizontal row) represents a global summary of the distribution of feature importance over the validation set. Features with asterisk in their name represent the expert ground truth (nodes with a positive value $t$ in the expert KG).  Generally, datapoints reflecting the GT for an accurate model such as this one, acknowledge properly attributed object part contributions to the classification in the right side of the plot, in pink.   }
\label{fig:shap_GM}
\end{figure}

In Figs. \ref{fig:shap_RB} and \ref{fig:shap_GM} %
we performed the SHAP analysis over the whole validation set. In practice it means that SHAP values were computed for each element of the validation set and plotted on the same graph. Then for each feature of the feature vector, in our case for each architectural element, we plot all SHAP values for this specific element found in the dataset, and we color them based on the feature value. They are plotted line-wise and each dot represents the feature value of a specific datapoint, i.e., image. High feature values (regarding the range they can take) are colored pink and low feature values in blue. Here, if an element is detected several times or with high detection confidence, it will be pink (blue for less detection confidence or less frequency). Then, horizontally, are shown the SHAP values, where high (absolute) values have high impact on the prediction\footnote{See tutorial \href{https://christophm.github.io/interpretable-ml-book/shap.html}{https://christophm.github.io/interpretable-ml-book/shap.html} and SHAP source code in \href{https://github.com/slundberg/shap}{https://github.com/slundberg/shap}. %
}.

If we compare the SHAP plots with the KG, here we do not observe any large amount of \textit{outliers} or datapoints not coinciding with the domain expert KG acting as ground truth (in Fig. \ref{fig:SAG_and_KG_projection} right). We now need to be able to use this information automatically. %
Pink and blue (high and low) values of datapoint features can appear both in right and left sides of the plots, meaning its value can contribute towards predicting the considered class or not, respectively. However, in our case, only pink datapoints being on the positive (right) side of SHAP plot represent the \textit{correct} behaviour if such element is also present in the KG. In that case, their feature value loss will not be penalized during training, as they match the expert KG (considered as GT). The rest of datapoints' SHAP values (blue in right side, pink and blue in left side) will be used by SHAP-Backprop to correct the predicted object class.  

An example of computation of SHAP values on a single feature vector is in Table \ref{tab:SHAP_analysis_example}. On the right there is the feature vector, and on the left the SHAP values for each object class for each object part. We highlighted in green positive values and in red negative values.

\subsubsection{SAG: SHAP Attribution Graph 
to compute an XAI loss and explainability metric}
\label{sec:SAG_and_l_shap}

By measuring how interpretable our model is, in the form of a KG, we want to be able to tell if the decision process of our model is similar to how an expert mentally organizes its knowledge. As highlighted in the previous section, thanks to SHAP we can see how each feature value impacts the predicted macro label and thus, how each part of an object class impacts the predicted label. Based on this, we can create a \textit{SHAP attribution graph (SAG)}. In this graph, the nodes are the object (macro) labels, and the parts are linked to a macro label if according to the SHAP algorithm, it played a contribution role toward predicting this label.

Building the SAG is a two step process. First we extract the feature vector representing the attributes detected (float values). Thanks to the detection model we get the predicted label from it. Feature vectors are the output of the aggregation function that are fed to the classification module.

Using as hyperparameter a threshold $s$\footnote{Default thresholds used in our case for detection were $s$ = 0.05 for both Faster R-CNN and RetinaNet, as they showed to  work best %
for numerical stability.}  on each feature value, we identify which architectural element we have truly detected in the image. Then, using the SHAP values computed for this feature vector, we create a SAG per image in the test set, %
and thus we link together feature values and predicted label probabilities inside the SAG. This way we use SHAP to analyse the output for all classes, not only the predicted one:
 
\begin{itemize}
 \item Having a \textit{positive SHAP} value means the detected feature contributes to predicting this label, given a trained classifier and an image. We thus add to the SAG such edge representing a \textit{present feature contribution}. 

 \item Having a \textit{negative SHAP} value and a feature value below the threshold $s$ means that this element is considered typical of this label and its absence is detrimental to the prediction. As such, we can link the object label and the part label in the SAG, as a \textit{lacking feature contribution}.
\end{itemize}
 
\textit{Lacking} feature contributions or \textit{present} feature contributions in the form of edges in the expert KG are obtained through a projection of the SHAP attribution graph, obtained from a SHAP analysis of the trained model, on the theoretical expert KG.  

An example of SAG for the architectural style classification problem is in Fig. \ref{fig:SAG_and_KG_projection}.
 
\textit{M, R, G, B} means \textit{Hispanic-Muslim, Renaissance, Gothic} and \textit{Baroque}, respectively. The pseudo code to generate the SAG can be found in Algorithm \ref{alg:sag}.

\begin{algorithm}
\caption{Computes the SHAP attribution graph (SAG) for a given inference sample.}
\begin{algorithmic}%
\REQUIRE feature\_vector, shap\_values, %
Classes, Parts, part detected threshold $s$
\STATE $SAG \leftarrow ${\{\}}
\FOR{class in Classes}
    \STATE $local\_shap \leftarrow shap\_values[class]$
    \FOR{part in Parts}
        \STATE $feature\_val \leftarrow feature\_vector[part]$
        \STATE $shap\_val \leftarrow local\_shap[part]$
        \IF{feature\_val $ > s$}
            \IF{shap\_val $ > 0$}
                \STATE ADD (part, object) edge to $SAG$
            \ENDIF
        \ELSE
            \IF{shap\_val $ < 0$}
                \STATE ADD (part, object) edge to $SAG$
            \ENDIF
        \ENDIF
    \ENDFOR
\ENDFOR
\RETURN $SAG$
\end{algorithmic}
\label{alg:sag}
\end{algorithm}

In practice this allow us to have an empirical attribution graph, the SAG (built at inference time), and a theoretical attribution graph, the KG (representing prior knowledge). We can then compare both of them.

\subsubsection{%
SHAP-Backprop to penalize misalignment with an expert KG}%
\label{sec:shapbackprop}

In order to improve performance and interpretability of the model, we hypothesize that incorporating the SHAP analysis during the training process can be useful to fuse explainable information and improve interpretability.

The underlying idea is that SHAP helps us understand on a local level what features are contributing toward what class. In our case, SHAP links  elements to a label by telling us if it contributed toward or against this label, and how strongly. Besides this analysis, we have the KG that is embedding basically the same information. 
We can see the KG as a set of expected attributions, i.e., if an element is linked to the object label in the KG, it should contribute to it, otherwise it should contribute against. 

Given these two facts, we can compare real attribution via SHAP analysis, that gives us the empirical attribution graph, with the theoretical attribution found in the KG. If there is a discrepancy, then we want to penalize this feature, either in the classification process or in the detection process. %

Misattribution, which happens when a feature attribution is unexpected or absent, can stem from various origins. One would be a recurrent misdetection inside the dataset. As such, penalizing misattribution at detection time could help us correct those.  Penalizing the classification process could be considered as well, but has not been done in this work.  

\begin{figure}[htbp!]%
    \centering
    \includegraphics[width=17cm%
    ]{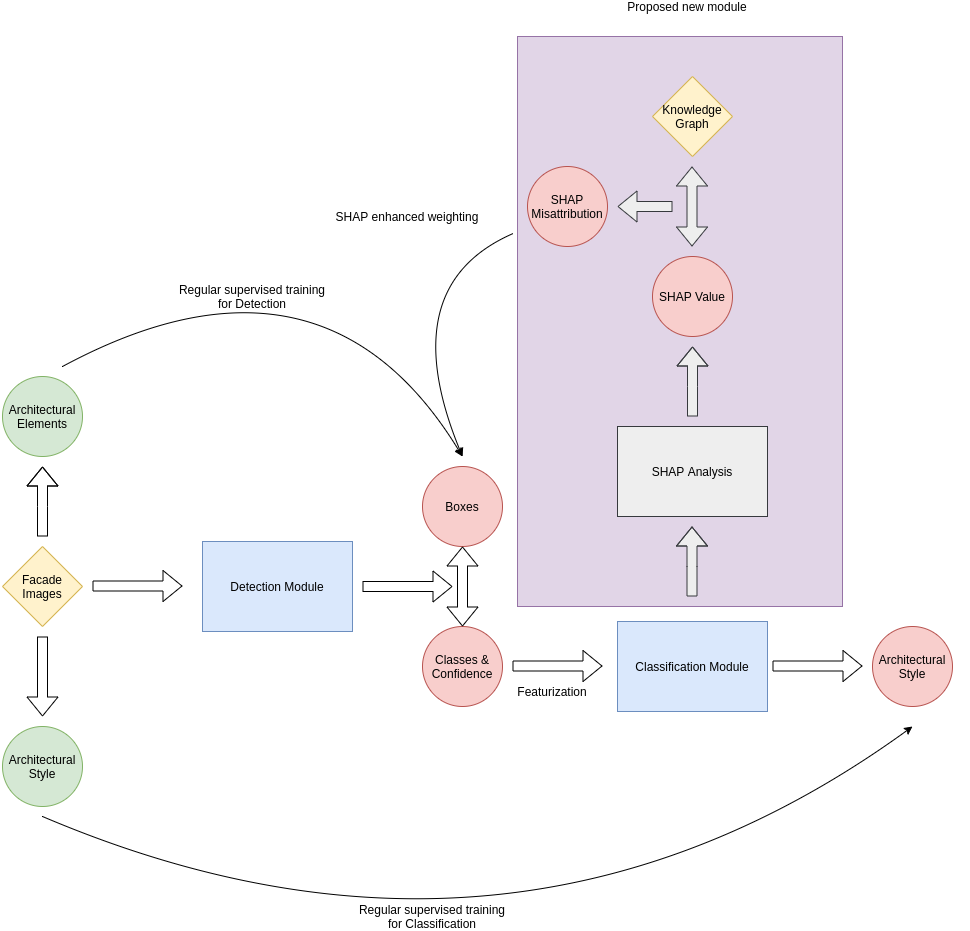}
    \caption{Diagram of proposed X-NeSyL methodology focused on the new  SHAP-Backprop training procedure (in purple). In yellow the input data, in green the ground truth elements, in red predicted values (including classes and bounding boxes), and in blue trainable modules.   
    Red blocks are the output of the various algorithms and gray blocks are untrainable elements. The purple block is the proposed X-NeSyL procedure. Thin arrows link together outputs of trainable modules along with reference elements used for backpropagation. Initially two thin arrows exist, one between the object (macro) label and the output of the classification model, and one between elements (object parts) and the output of the detection module. We add a third arrow (i.e., feedback signal) between the SHAP misattribution and the output of the object part detector as the new SHAP-penalized loss is leveraging this misattribution to penalize misalignment with the expert.  }
    \label{fig:new_pipeline}
\end{figure}

A schema of this approach is presented in Fig. \ref{fig:new_pipeline}. A new misattribution loss requires an intertwined training of the classification model with the detection model to compute the SHAP analysis; however the extra required training time in practice is not a big issue. Indeed, in the initial training protocol, one would fully train the detection and then the classification. Here, we have to train the classification at each detection epoch. We expect the explainability metric, i.e., the SHAP GED between the  KG and the SAG to increase thanks to this SHAP signal backpropagation.

\subsubsection{\cal{L}\textsubscript{SHAP}:
A new loss function based on SHAP values and a misattribution function $\beta$}
Let $N$ be the number of training examples and let $I = (I_1,I_2,...,I_N)$ be the training image examples. 

Let $D$ be the detector function such as: 
\begin{equation}
    D(I_i) = (BB_i, Conf_i, Class_i),
\end{equation}
where $BB_i = (BB_1,BB_2,...,BB_{N_i})$ are the bounding boxes detected by D, $Conf_i$ the confidence associated to each predicted box, and $Class_i$ the predicted class of each box. The associated ground truth label is used for standard backpropagation, but we will not need it for the weighting process.

Faster R-CNN \cite{ren2015faster} object detector uses a two terms loss: 

\begin{equation}
\mathcal{L}_{detector} = \mathcal{L}_{RPN} + \mathcal{L}_{ROI},
\end{equation} %
where $\mathcal{L}_{ROI}$ is the loss corresponding to predicting the region of interest and is dependant on the class predicted for each BB. It is computed at the output class level, whereas $\mathcal{L}_{RPN}$ is the loss function from the region proposal network, and it is computed at the anchor level\footnote{Anchors are a set of predefined bounding boxes of a certain height and width. These boxes are defined to capture the scale and aspect ratio of specific objects. Height and width of anchors are a hyperparameter chosen when initializing the network.}. We use a weighted 
$\mathcal{L}_{ROI}$, since the SHAP information is computed at the output class level and not the anchor level. We can write the loss as the sum of the losses for each image, and within an image, for each BB:

\begin{equation}
    \mathcal{L}_{ROI} = \Sigma_{i=1}^{N}\Sigma_{k=1}^{N_i}\mathcal{L}(I_i,(BB_k,Conf_k,Class_k)),
\end{equation}
where $i$ is the index of the considered image, $k$ the index of a BB predicted within that image,  and $\mathcal{L}$ is a weighted sum of cross entropy and smooth $L_1$ loss. 

We now introduce the SHAP values, which are used as a constraining mechanism of our classifier model to be aligned with expert knowledge provided in the KG. SHAP values are computed after training the classification model. 

 Let $S = (S_1,S_2,...,S_N)$ with $Si = (S_{i,1},...,S_{i,m}) $, with $m$ the number of different object (macro) labels, and let $S_{i,k} = (S_{i,k,1},S_{i,k,2},...,S_{i,k,l})$ be the SHAP values for each training example $i$, where $k$ is the macro object label, and $l$ is the detected part. Each SHAP value is thus of size $l$, with $l$ being the number of parts in the model. Furthermore, due to the nature of the output of the classification model, which are probabilities, and the way SHAP values are computed, they are bounded %
 to be a real number in $[-1, 1]$.

The KG was already modeled as an attribution graph and corresponding matrix (%
in order to compute the embedding out of the KG) in Section \ref{sec:SAG_and_l_shap}, and we will be using the same notation.

\textbf{Introducing the misattribution function $\beta$}

To introduce SHAP-Backprop into our training, we first need to be able to compare the SHAP values with a ground truth, which here is represented by the expert KG. We thus introduce the misattribution function to assess the level of alignment of the feature attribution SHAP values with the expert KG. %

The goal of the misattribution function is to quantitatively compare the SHAP values computed for the training examples with the KG. For that we assume the SHAP values are computed for all feature vectors. %
A misattribution value is then computed for each feature value of each feature vector. 
Before considering the definition of misattribution function, we can distinguish two cases when comparing these two elements, depending on the feature values observed: 

A) The feature value considered is higher than a given hyperparameter $v$, i.e., the positive case. $v$ symbolizes the value above which we consider a part is detected in our sample image. In our case $v = 0$. 

B) The feature value is lower or equal to $v$: in this case we assume there is no detected part, i.e., the negative case.

\textit{Case A}: In the first case, given the KG, for a SHAP attribution to be coherent, it should have the same sign as the KG. If it is the case, the misattribution is 0, i.e., there is no correction to be made and backpropagate. Otherwise, if it has opposite sign, the misattribution will depend on the SHAP value. In particular, it will be proportional to the absolute value of the SHAP attribution. We thus propose the following misattribution function:

\begin{equation}
    \beta(S,KG,i,k,j) = (-KG_{k,j} \times S_{i,k,j})^+,
\end{equation}
where $i$ is the index of the considered image, $k$ is the index of a given object (macro) label and $j$ is the index of a given (object) part. This way $KG_{k,j}$ correspond to the edge value between the macro label $k$ and the part $j$, where $(.)^+$ is the positive part of a real number.
$KG_{k,j} = \pm 1$, and thus $\beta \in [0,1]$ due to the bounding of the SHAP values. 
The detector output feature values are bounded due to the nature of the classification output which is in [0,1]. %
However, SHAP values %
are naturally bounded in [-1,1] \cite{lundberg2017unified}.

\textit{Case B}: Since we choose $v$ to be $0$ and $v$ has only real values, if $v=0$, we therefore should not backpropagate any error through the loss function, since no BB is detected for the object part.

Given the prior information in the KG, the posterior information (SHAP values post-training), and a way to compare them (attribution function $beta$), we suggest two new versions of a weighted loss, $\mathcal{L}_{SHAP}$, that will substitute the former ROI loss.

\textit{- Bounding Box-level weighting of the $\mathcal{L}_{SHAP}$ loss}

This first weighted loss is at the bounding box (BB) level, meaning each BB will be weighted individually based on its label and the associated SHAP value. We propose the following loss:
\begin{equation}
    \mathcal{L}_{SHAP} = \Sigma_{i=1}^{N}\Sigma_{k=1}^{N_i}\alpha_{BBox}(S,KG,i,C_i,Class_k)\mathcal{L}(I_i,(BB_k,Conf_k,Class_k)),
\end{equation}
where $N_i$ is the number of BBs predicted in image $I_i$, and $C = (C_1,...,C_N)$ the ground truth (GT) labels for instance images $I = (I_1,I_2,...,I_N)$.
We propose two possible loss weighting options, depending on $h$, a balancing hyperparameter (equal to $1$ in our experiments), that can be linear:
\begin{equation}
    \alpha_{BBox}(S,KG,i,C_i,Class_k) = h*\beta(S,KG,i,C_i,Class_k) + 1
\end{equation}
or exponential:
\begin{equation}
    \alpha_{BBox}(S,KG,i,C_i,Class_k) = e^{h*\beta(S,KG,i,C_i,Class_k)},
\end{equation}
with $i$ the index of the considered image, $C_i$ its associated class, $Class_k$ the considered part class, $KG$ the KG and $S$ the SHAP values.  
Either way, if $\alpha$ is equal to $1$, then the misattribution is $0$ in order to maintain the value of the original loss function. Thus, $\alpha \in [1, \infty)$: the larger the misattribution, the larger the penalization.   

\textit{- Instance-level weighting of the $\mathcal{L}_{SHAP}$ loss}

This second weighted loss is at the instance-level, meaning we are weighting all the BBs for a given dataset instance  (i.e., image)   with the same value:
\begin{equation}
   \mathcal{L}_{SHAP} = \Sigma_{i=1}^{N}\alpha_{instance}(S,KG,i,C_i)\Sigma_{k=1}^{N_i}\mathcal{L}(I_i,(BB_k,Conf_k,Class_k)),
\end{equation}
where 
\begin{equation}
\alpha_{instance}(S,KG,i,k) = max_{j \in [1,l]}(\alpha_{BBox}(S,KG,i,k,j))
\end{equation}
i.e., the instance-level weighting of the loss function considers the max BBox misattribution function value. Just as the BB level weighting, the %
aggregation of terms in the misattribution function can either be linear or exponential.

\section{X-NeSyL methodology Evaluation: SHAP GED metric to report model explainability for end-user and domain expert audiences} %
\label{sec:shapged}

Detection and classification modules of EXPLANet use mAP and Accuracy, respectively, as standard evaluation metric. In order to evaluate explainability of the model in terms of alignment with the KG, we propose the use of the SHAP Graph Edit Distance (SHAP GED) at test time.  This metric %
has a well defined target audience: the end-user (in our case, of a citizen science application) and domain experts (art historians), i.e., users with non-technical background necessarily.

Even if the SAG above can be computed for any set of theoretical and empirical feature attribution sets, we are interested in using the GT KG in order to compute a explainability score on a test set.

The simplest way to compare two graphs is applying the GED \cite{sanfeliu1983distance}. Using straight up the GED between a KG and the SAG does not work very well, since the number of object parts (architectural elements in our case) detected vary too much from an image to another. What we do is to compare the SAG to the projection of the KG given the nodes present in the SAG. More precisely, given a SAG, we compute a new graph from the KG, where we take the subgraph of the KG that only contains the nodes in the SAG. As, such they will have the same nodes, but with the potential addition of new edges.

\begin{figure}[htbp!]%
\centering
\begin{minipage}{.5\textwidth}
    \centering
    \includegraphics[width=\linewidth]{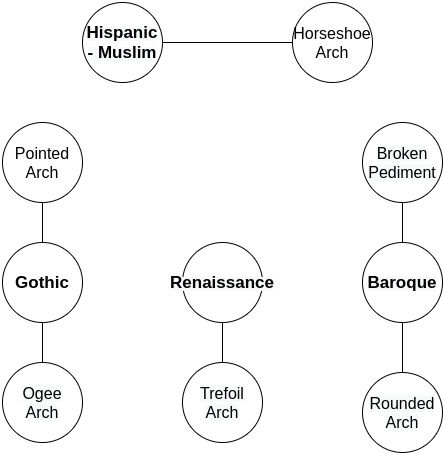}
    \label{fig:attribution_graph}
\end{minipage}%
\vline
\begin{minipage}{.5\textwidth}
    \centering
    \includegraphics[width=\linewidth]{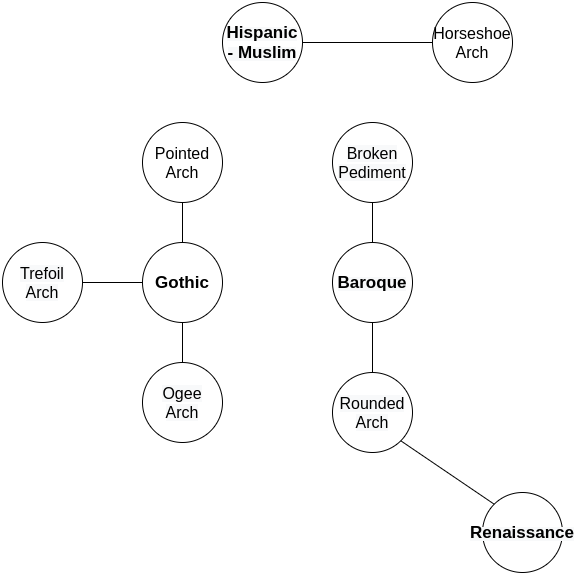}
    \label{fig:projected_graph}
\end{minipage}
    \caption{   Example of applying SHAP analysis after training an architectural style classifier based on an architectural element detector, and computing a SHAP attribution graph (SAG, on the left). Given the nodes present in such SAG, the projection on the KG is performed (on the right) to get the non matching edges (and the graph Edit distance, GED) and compute the misattribution function $\beta$. The latter penalizes the model for assigning feature attribution that conflicts with the attribution of features an expert would assign (represented in Fig. \ref{fig:taxonomy} expert KG). %
     }
    
    \label{fig:SAG_and_KG_projection}
\end{figure}

An example of such projection can be seen in Figure \ref{fig:SAG_and_KG_projection} (right). This way, the projection serves to only compute the relevant information given a specific image.

\begin{table}[htbp!]
    \centering
    \small
    \setlength\extrarowheight{3pt}
    \begin{tabular*}{1\textwidth}{@{\extracolsep{\fill}}l|rrrr|r@{}}
        \hline
         & \textbf{Hisp.-Muslim} & \textbf{Gothic} & \textbf{Renaissance} & \textbf{Baroque} & \textbf{Feature vector} \\
        \hline
        Horseshoe arch          & \cellcolor{red!30}-0.16 & 0     & \cellcolor{green!25}0.08  & \cellcolor{green!10}0.03  & 0 \\
        Lobed arch              & 0     & 0     & 0     & 0     & 0 \\
        Flat arch               & 0     & 0     & 0     & 0     & 0 \\
        Pointed arch            & 0     & \cellcolor{red!30}-0.15 & \cellcolor{green!25}0.07  & \cellcolor{green!10}0.04  & 0 \\
        Ogee arch               & 0     & \cellcolor{red!15}-0.08 & \cellcolor{green!10}0.01  & 0     & 0 \\
        Trefoil arch            & 0     & 0     & \cellcolor{green!10}0.04  & 0     & \cellcolor{green!25}0.2 \\
        Triangular pediment     & 0     & 0     & 0     & \cellcolor{green!20}0.06  & 0 \\
        Segmental pediment      & 0     & 0     & 0     & 0     & 0 \\
        Serliana                & 0     & 0     & 0     & 0     & 0 \\
        Rounded arch            & 0     & 0     & 0     & \cellcolor{green!10}0.03  & \cellcolor{green!50}1.35 \\
        Lintelled doorway       & 0     & 0     & 0     & 0     & 0 \\
        Porthole                & 0     & 0     & 0     & 0     & 0 \\
        Broken pediment         & 0     & 0     & \cellcolor{green!35}0.14  & \cellcolor{red!30}-0.16 & 0 \\
        Solomonic column        & 0     & 0     & \cellcolor{green!15}0.04  & 0     & 0 \\
        \hline
    \end{tabular*}
    \caption{Feature vector of a sample image and its SHAP analysis used for the construction of the SAG in Fig. \ref{fig:SAG_and_KG_projection}, according to Algorithm \ref{alg:sag}.}%
    \label{tab:SHAP_analysis_example}
\end{table}

Once SHAP-Backprop procedure penalizes the missalignment of object parts with those of the KG (detailed in next section), we will use the SAG to compute the SHAP GED between the SAG and its projection in the KG. This procedure basically translates into counting the number of "wrong" edges in the SAG given the reference KG, i.e., the %
object parts that should not be present in this data point, given the predicted object label. %

After detailing all necessary components to run the full pipeline of X-NeSyL methodology, together with an evaluation metric that facilitates the full process evaluation, we are in place to set up an experimental study to validate each component. It is worth noting that each component can be adapted to each use case. Next section experiments will demonstrate, with a real life dataset, how X-NeSyL methodology can facilitate the learning of explainable features, by fusing the information from deep and symbolic representations.

\section{%
MonuMAI Case Study: Classifying monument facades architectonic styles}%
\label{sec:experiments}

In order to evaluate the X-NeSyL methodology, and all inherent components including the detection and classification module within the proposed EXPLANet architecture, as well as the XAI training procedure, we perform two main studies. In the first study, we evaluate the full SHAP-Backprop training mechanism by testing the detection module followed by the classification one. In other words, we test and assess the full EXPLANet architecture. 
In the second study, as an ablation study to assess the influence of the detector's accuracy on the overall part-based classifier, we evaluate the detection module of EXPLANet model with two different detection models,  Fast-RCNN \cite{ren2015faster} and RetinaNet \cite{lin2017focal}. 

In both evaluation studies, we used two datasets, MonuMAI and PASCAL-Part. For simplicity, we focus on MonuMAI dataset mainly in this section, while additional results for PASCAL-Part can be seen in the Appendix. %
For the remainder of the paper, we will use elements and object parts interchangeably, and macro object labels will be used to refer to the classification labels, i.e., the style of the macro object. 

\subsection{Experimental setup}

To evaluate the classification performance  we use the standard  accuracy metric (equation \ref{ec:precrecf1}).

\begin{align}
    \label{ec:precrecf1}
    \begin{split}
        accuracy &= \frac{\#correct\ predictions}{\#total\ predictions} \\
    \end{split}
\end{align},
where $\#$ represents the number for correct and total predictions. 
To evaluate the detection performance, we use the standard metric mean average precision \textit{mAP} (Eq. \ref{ec:map}).  

\begin{align}
    \label{ec:map}
    mAP &= \frac{\sum_{i=1}^{K} AP_{i}}{K}     &   AP_{i} &= \frac{1}{10} \sum_{r\in[0.5,...,0.95]} \int_{0}^{1}p(r)dr
\end{align}

where given $K$ categories of elements, $p$ precision and r $recall$ define $p(r)$ as the area under the interpolated precision-recall curve for class $i$.

We initialized Faster R-CNN \cite{ren2015faster} and RetinaNet with the pre-trained weights on MS-COCO \cite{lin2014microsoft} then fine-tuned both detection architectures on the target datasets, i.e., MonuMAI or PASCAL-Part.
The last two layers of Faster R-CNN were fine-tuned on the target dataset. As optimization method, we used Stochastic Gradient Descent (SGD) with learning rate of $0.0003$ and a momentum of $0.1$.  We use \textit{Faster R-CNN} implementation provided by  native PyTorch's \textit{torchvision} library. Concretely, Faster R-CNN is implemented using PyTorch packages \textit{\textit{torch} 1.7.0, \textit{torchvision}  0.8.1}. The classifier after the detector uses Tensorflow 1.15 (\textit{tensorflow-gpu 1.15.0}).  

For the classification module, we also fine-tuned the two layer MLP with 11 intermediate neurons. We used the Adam \cite{kingma2014adam} optimizer provided by Keras. 

To perform an ablation study on the element or part-based detector,  the original dataset is split into three categories (train, validation and test), following a 60/20/20 split. Reported results are computed on the test set. 

The compositional part-based object classification with RetinaNet is trained in two phases. First the detection is trained by finetuning a RetinaNet-50 pretrained on MS COCO. We use Adam optimizer with starting learning rate (LR) of $0.00001$ and a scheduler of learning rate to reduce on plateau with a patience of 4\footnote{Patience is the number of epochs taken into account for the scheduler to %
decide the network converged. Here, the last four.}. We train this way for 50 epochs. Then we freeze the whole detection weights and train only the classification. We use Adam optimizer with starting LR= $0.001$ and a scheduler of LR to reduce on plateau with a patience of 4. We train this way for 25 epochs.

Even if our objective was having a fully end-to-end training, the need for a quite different LR between the detection and classification modules led us to train separately for convenience, at the moment.

\subsection{EXPLANet model analysis}
In order to assess the advantages of EXPLANet, we consider  three   baselines: 1) MonuNet \cite{lamas2020monumai}: the architecture proposed with MonuMAI dataset, designed as a compressed architecture that is able to run in embedded devices such as smartphones\footnote{Since MonuMaiKET detector and MonuNet classifier are not connected, MonuNet does not provide object detection.}.  
2) A simple object classifier based on vanilla ResNet-101 \cite{he2016deep}. MonuNet is a different classification architecture to ResNet, it uses residual and inception blocks but with a more static architecture than EXPLANet that does not %
allow modifications or is not meant to be scalable. 
 
3) A KG Deterministic classifier baseline, based on the same pipeline as EXPLANet, but using solely the expert KG. It first computes the detection scores and uses the same aggregation function (Eq. 1) as EXPLANet, but then the classification is done by summing for each class the contribution of all related object-parts that are typical of this object class, according to the expert KG. The resulting vector is normalized (to resize the output between 0 and 1 so it sums to 1) and an \textit{arg max} is applied to decide the style, without any training. 
For an image $I$, the KG Deterministic classifier ($DC_{KG}$) is defined as follow:
\begin{equation}
    DC_{KG}(I) = argmax_{k\in[1,m]}(confidence_k(I)) 
\end{equation}
where
\begin{equation}
    confidence_k(I) = \sum_{j \in [1,l]}\frac{1+KG_{k,j}}{2}v_j(I) \\
\end{equation}
with $v$ the image descriptor based on the detected object parts defined in Eq. 1, $KG$ the knowledge graph defined in Eg. 5, and $k \in [1,m]$ the object index out of $m$ classes and $j \in [1,l]$ the object-part index.

The results of EXPLANet classification model based on Faster R-CNN and RetinaNet detector backbones, together with these baseline classification networks are shown in Table \ref{tab:result} for MonuMAI dataset. %

\begin{table}[htbp!]
    \centering
    \begin{tabular}{c|c|c|c}
    & \textbf{mAP} & \textbf{Accuracy} & \textbf{SHAP GED}  \\
    & (detection)  & (classification)  & (interpretability) \\
    \hline
    EXPLANet using Faster R-CNN backbone detector &  $42.5$ & $86.8$ & $0.93$\\ 
    \hline
    EXPLANet using RetinaNet backbone detector & \textbf{49.5} & \textbf{90.4} & \textbf{0.86}\\ 
    \hline
    ResNet-101 classifier baseline &  N/A & $84.9$ & N/A\\ 
    \hline
    MonuNet classifier baseline &  N/A & $83.11$ & N/A \\ 
    \hline
     
    KG Deterministic classifier baseline  &  N/A & $54.79$ & N/A \\ 
     
    \end{tabular}
    \caption{Compositional vs \textit{direct} (traditional) classification: Results of two backbone variants of EXPLANet (using object detector Faster R-CNN and RetinaNet) on MonuMAI dataset, and comparison with embedded version of the baseline model MonuNet, a vanilla classifier baseline with ResNet,   and an expert KG-based deterministic (non-trained) classifier. In this baseline, the same results are obtained using majority voting based on the detection output of both object detectors, Faster R-CNN and RetinaNet (i.e., despite concrete examples having different confidence scores, the total right guesses are the same and result in the same accuracy).       %
     EXPLANet versions use the standard procedure, no SHAP-Backprop, to demonstrate that a part-based compositional classifier improves over a straight classifier (i.e., without a connected object-part detector as input).}%
    \label{tab:result}
\end{table}

Both versions of EXPLANet %
outperform the three baselines. Here results are shown with a Faster R-CNN baseline, but accuracy remains the same for RetinaNet, given that the majority voting on different attribute detection confidence scores results in the same final accuracy.  %
In average, RetinaNet produces a detection and classification of better quality.

MonuNet \cite{lamas2020monumai}, the baseline provided by MonuMAI dataset authors, is an architecture designed for being used in mobile devices in real time. %
Because of its compressed design that targets embedded systems, its performance is not fully comparable with EXPLANet. However, we report it for reference as, to the best of our knowledge, it is the only previous model trained on the novel MonuMAI dataset to date. %

Due to the naturally simpler nature of RetinaNet, the latter is faster to train than Faster R-CNN\footnote{We use the RetinaNet 100\% PyTorch   1.4 (torch 1.7.0, torchvision  0.8.1)   implementation from \href{https://github.com/yhenon/pytorch-retinanet}{https://github.com/yhenon/pytorch-retinanet}. Ease of use stems from the fact that if we wanted to modify the aggregation function, whether its analytical form or at the end of the detector, at which we should attach the classifier, it would be much simpler.}.

\begin{figure}[htbp!]
\centering
\begin{minipage}{.59\textwidth}
    \centering
    \includegraphics[width=\linewidth]{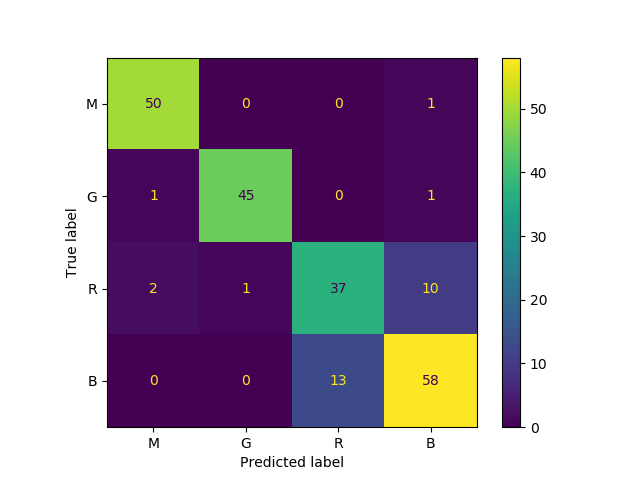}
    \label{fig:EXPLANet_MonuMAI}
\end{minipage}%
\begin{minipage}{.59\textwidth}
    \centering
    \includegraphics[width=\linewidth]{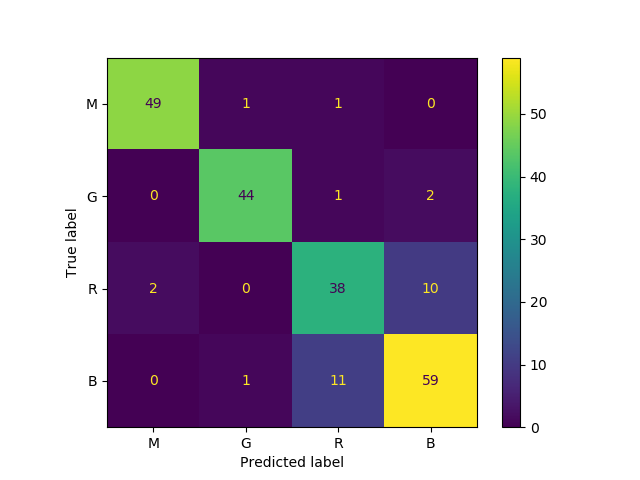}
    \label{fig:RetinaNet_MonuMAI}
\end{minipage}
\caption{Confusion matrix of EXPLANet using Faster R-CNN (left) and EXPLANet using RetinaNet (right) on MonuMAI dataset for each architectural style class.}
\label{fig:confmatrices}
\end{figure}

We can see the confusion matrix computed on MonuMAI for EXPLANet, using both Faster R-CNN and RetinaNet object detectors as backbones, in Fig. \ref{fig:confmatrices}.
Although a better detector (better mAP) could be intuitive to encourage a better GED, it is not expected, because the mAP evaluates the spatial location and the presence or not of a descriptor (object part), while the GED evaluates just the presence. %
Moreover, %
having no correlation among mAP and GED is reasonable, specially because mAP evaluates only one part of the model (detection), and thus it makes more sense that accuracy correlates with SHAP GED, as our results show for MonuMAI. %

The object part detector module of MonuNet baseline, i.e., MonuMAIKET detector (based on Faster R-CNN detector -using ResNet101- as backbone) reaches slightly higher performance. We assume this minor difference due to the different TensorFlow and PyTorch default implementations of Faster R-CNN's inherent ResNet module versions in MonuNet and EXPLANet, respectively). %

Interpretability-wise, EXPLANet with RetinaNet approach outperforms EXPLANet with Faster R-CNN. This probably stems from the object parts aggregation functions that are slightly different. In the Faster R-CNN version of EXPLANet, only the %
probability for the highest scoring label is kept, whereas in the case of using RetinaNet as part detector, the latter aggregates over all the scores for each example, %
with the sum function. This way RetinaNet is probably more robust to low score features as it %
always %
contemplates the several labels detected for each example. %
 
At the moment minimal work is done to process the aggregating vector (i.e., vector accumulating the probabilities of each bounding box), but that could be explored in the future.

We observe that the more accurate the detector is, the more accurate the classification model is, and the more interpretable (lower SHAP GED) it becomes. However, future work should further investigate more complex tasks and datasets (as appendix will show on PASCAL-Part experiments), since not having annotations with sufficient quality %
can be detrimental to interpretability, and reinforce the interpretability-explainability trade-off.

The takeaway message from this experiment is that overall, both part-based models outperform the regular classification for MonuMAI, which means that detecting prominent and relevant object parts helps the classifier (a MLP) focusing on the important elements to draw a decision. In other words, detecting object parts helps the model putting attention on the parts of the image that matter, and making the prediction more interpretable. Furthermore, since ResNet is highly non interpretable on its own, tabular data is much easier to process for XAI methods.

\subsection{SHAP-Backprop training analysis}
Once assessed the EXPLANet architecture as a whole, and once performed an ablation study with respect to dependency on the object detector, we asses the different ways of weighting the training procedure penalization.

Table \ref{tab:shap_backprop_results_monumai} %
displays the computed results of X-NeSyL methodology SHAP-Backprop method on %
\textit{MonuMAI} \cite{lamas2020monumai} dataset (see additional results for PASCAL-Part \cite{everingham2010pascal} in Appendix). %
In bold are the best results for each metric, and in italics the second best. We tested on what we call \textit{Standard procedure}, which is the typical pipeline of training first the detection module and then the classification module in two different steps, sequentially, without SHAP-Backprop nor any other interference. 

The four other cases are methods computed with SHAP-Backprop. At each epoch in the detection phase, we train a classifier and use it to compute the SHAP values. These are then used to weight the detection loss with the misattribution function presented in the previous subsection.

\begin{table}[htbp!]
    \centering
    \begin{tabular}{c|c|c|c}
        & \textbf{mAP}  & \textbf{Accuracy} & \textbf{SHAP GED}\\
        & (detection) & (classification) & (interpretability) \\
        \hline
        Standard procedure (baseline, no SHAP-Backprop) &  $\textit{42.5}$ & $86.8$ & $0.93$\\
        \hline
        Linear BBox-level weighting & $42.4$ & $86.3$ & $\textit{0.69}$\\
        \hline
        Exponential BBox-level weighting & $42.3$ & $\textbf{88.6}$ & $1.16$  \\
        \hline
        Linear instance-level weighting & $41.9$ & $87.2$ & $\textbf{0.55}$\\
        \hline
        Exponential instance-level weighting & $\textbf{44.2}$ & $\textbf{88.6}$ & $1.09$  \\
    \end{tabular}
    \caption{Impact of the new SHAP-Backprop training procedure on the mAP (detection), accuracy (classification) and SHAP GED (interpretability). Results on MonuMAI dataset with EXPLANet architecture   show a positive improvement on interpretability by using SHAP-Backprop with linear instance-level weighting.    SHAP GED is computed between the SAG and its projection in the expert KG. (\textit{Standard procedure} means a sequential typical pipeline of 1) train detector, 2) train classifier with no SHAP-Backprop).}%
    \label{tab:shap_backprop_results_monumai}
\end{table}

Table \ref{tab:shap_backprop_results_monumai}. Applying SHAP-Backprop has %
little effect on accuracy and mAP. Nonetheless, all but the linear BBox-level weighting increase the classifier accuracy around 1-2\%. %
These instabilities could probably be stabilized with further domain specific fine-tuning. Furthermore, we have to take into account the stochastic nature of the training process, since the SHAP value computation is %
approximated %
using a random subset of reference examples, to be more computationally efficient.%
 
On the other side, in terms of interpretability, we do have a more sensible
improvement, (reducing SHAP GED from 0.93 to 0.55) in the case of linear instance weighting. 

 The takeaway of this experiment is that there are other ways to improve interpretability than a good detector. For instance, using SHAP-Backprop. In particular, its linear weighting scheme here is stronger than exponential. %
 
The gain obtained in both dimensions is large enough to conclude that the X-NeSyL methodology helps improving interpretability in terms of SHAP GED to the expert KG.

\subsection{%
Lessons learned}
\label{sec:lessonsLearnedAndFutureWork}

In order to create a NeSy deep learning model that is explainable, we chose the expert as target audience of the model output explanation. We then proposed a training procedure and metric to qualitatively asses domain expert based explanations. Although further explainability metrics beyond SHAP GED could be studied depending on the audience of the explanation, we showed explainability, in terms of alignment of the conceptual match with the domain expert KG is increased.

The X-NeSyL methodology with pluggable components is meant to be a generic and versatile one, i.e., a template architecture that can be adapted and customized to each use case: the symbolic component for knowledge representation, the neural component based on a compositional architecture such as EXPLANet, and the XAI-informed misattribution procedure to be applied during training.
Furthermore, our experiments verified our hypotheses:
\begin{enumerate}
    \item Overall, X-NeSyL methodology brings explainability in the fusion of deep learning representations with domain expert knowledge. X-NeSyL did have the expected effect on the MonuMAI dataset, and SHAP-Backprop %
    improves the %
    explainability metric (SHAP GED) on the model. %
    \item The intermediate learned representation of EXPLANet allows to remove noisy information.   Learning to classify by detecting object parts first shows to be more interpretable can also improve precision. For instance, some monuments are defined by a single element (e.g., \textit{broken pediment}), to which the system learns to give an importance greater than the others.  

    \item %

Not always more accurate means more interpretable. However, we showed with the MonuMAI use case that it it possible that the more accurate EXPLANet classifier model is, the more interpretable (lower SHAP GED) it becomes\footnote{As explained in the PASCAL-Part dataset experiments, this is due to a debility of PASCAL-Part dataset and its label names not being specifically designed for a compositional whole-part classification task (since they become noisy as they use the same object part label for semantically different concepts).}. Even if different error weighting schemes in SHAP-Backprop %
display this phenomenon is not always the case, we encourage future avenues of research to further explore this hypothesis on more complex tasks.
      
    \item Despite some weighting schemes confirming the interpretability-performance trade-off (some improve SHAP GED while worsening accuracy, and viceversa),   if the main objective is improving interpretability, we recommend using SHAP-Backprop with the linear instance-level weighting scheme. This is able to improve over both interpretability and performance.  %
    
    \item    Linear weighting yields better results than exponential due to the distribution of the misattribution function depending on the SHAP value. Exponential weighting focuses on examples that are extremely wrong, whereas the linear one gives a similar importance to all wrong examples. As SHAP GED does not take into account the SHAP value, but only whether there is a connection in the attribution graph, negating or correcting all examples similarly is the best approach when seeking explainability. %
Regarding bounding box aggregation function, instance-level works better than the BBox-level because the latter might slightly overfit each box, whereas instance-level gives a more average value to each image.

\end{enumerate}

As there is no consensus on how to measure explainability, especially because methods are achieving different goals, the lack of unification moved us to develop and contribute our own metric, SHAP GED, because as far as we know no other work explicitly incorporates the use of expert domain KGs in an image classification process to produce explanations for end-users and domain experts. %
We encourage researchers to put effort to further explore expert knowledge alignment models and develop richer metrics beyond this scope. 

When it comes to the semantic modelling of the KG, we explored the possibility of using the ontology OWL format, but compared to standard ontologies such as ArCo \cite{carriero2019arco} or the Google Knowledge Graph, our domain knowledge on architectonic styles is rather %
flat in terms of hierarchies of triples. We therefore did not need in this case the additional semantics modelling power of the Web Ontology Language (OWL) and limited ourselves to explain \textit{partOf} relationships. This permitted us to simplify the explanations and edge semantics. %
Future work should consider more complex semantic constraints natural of OWL format. 

Regarding the reproducibility or scalability of X-NeSyL methodology, manually constructing KGs for a given dataset in our case was not hard, given the scale of MonuMAI or PASCAL-Part datasets. For larger datasets where no domain expert KG is available, one debility of X-NeSyL methodology, in concretely of using KGs as symbolic knowledge representation, is the needs for the domain expert to design the KG. This may require, if experts are limited, or data is disperse and sparse, to previously recur to knowledge engineering tasks, among others, automatic knowledge base construction and datatype learning \cite{huitzil2018datil,diaz2017couch}, relation learning \cite{maruhashi2018learning}, link prediction \cite{getoor2005link,suchanek2019knowledge}, concept induction \cite{sarker2019efficient} or entity alignment \cite{Zhao20}.

\section{Conclusions and Future work} %
\label{sec:futureworks}

With the presented work we open up different research horizons and future avenues of work that we detail in this section.

We extensively considered %
what is one of the most crucial points to be addressed while developing XAI methods. %
Within the general needs for producing more trustworthy outputs, we tackled the challenge of fusion and alignment of deep learning representations with domain expert knowledge. To achieve this we proposed a new methodology, X-NeSyL, to fuse deep and symbolic representations thanks to an explainability feedback mechanism that facilitates the alignment of both deep and symbolic features. The part-based detection and classification, EXPLANet, and XAI-informed training procedure %
SHAP-Backprop leverage expert information in form of a knowledge graph. %
X-NeSyL could be seen as one way to attain explainable and theory-driven data science \cite{arrieta2020explainable}. 

We demonstrated the full pipeline of X-NeSyL methodology on MonuMAI and PASCAL-Part datasets, and the EXPLANet model with two variants of object detectors.
The fusion of learned representations of different nature through the addition of an XAI technique component facilitates the model to learn with a human expert in the loop.

X-NeSyL methodology was also validated through a contributed %
 audience-specific explainability metric, SHAP GED, that quantifies the alignment of the X-NeSyL methodology neural model (EXPLANet) with the symbolic representation of the expert knowledge. All models, datasets, training pipeline and metric of the showcased X-NeSyL methodology are available online\footnote{Models available online: \href{https://github.com/JulesSanchez/X-NeSyL}{https://github.com/JulesSanchez/X-NeSyL}}. %

This approach targeted compositional object recognition based on explaining the whole through the object-parts on deep architectures. However, other non compositional semantic properties of description logics could be further modelled in order to assess, and further constrain the level of alignment of a DL model with symbolic knowledge representing the expert.

Given the diverse contributions of this work, there is a broad set of options that can follow up to improve X-NeSyL methodology.
 In terms of evaluation of our work, the assessment was limited by the number of available datasets that contain part-based data, which is not large, since they must include a corresponding KG as well. %

The explainability metric may be refined, since the proposed vanilla version of SHAP GED might not take into account all explainable factors an expert would like to see reflected in a black box model explanation. Future work includes assessing the SHAP GED metric itself, as the most suitable graph comparison metric, and including more elaborated datasets with finer grained object-part labels. %

The ontology alignment with the deep model predictions can be refined in many ways. For instance, instead of using a simple KG, representing the expert knowledge in a rich ontology that incorporates extra axiomatic restrictions between elements, such as spatial relations or geometric constraints, could be useful to further improve SHAP-Backprop. %

One way to improve the model along these lines may be inducing 
spatial structure in the embedding space (e.g. with approaches such as \textit{ConvE} \cite{dettmers2018conve}, which uses CNNs on embeddings for link prediction). 
  Works along these lines could be explored to make possible that a KG would be learned %
(instead of being given to the DL model).   %

An actionable future work that could be very valuable for the XAI field is providing textual explanations of the output, since even limiting the model to describe the SAG could help build trust in the model output. %

To conclude, we invite researchers and domain experts to be part of the XAI debate and contribute to democratize XAI, and to collaboratively design quantitative metrics and assessment methods %
aimed at developers, domain experts and end-users, as target audiences of DL model explanations.

\section*{Acknowledgement(s)}
This research was funded by the French ANRT (Association Nationale Recherche Technologie - ANRT) industrial Cifre PhD contract with SEGULA Technologies, the Andalusian Excellence project P18-FR-4961 and the Spanish National Project
PID2020-119478GB-I00. The paper has also been partially supported by the Andalusian Excellence project P18-FR-4961. S. Tabik was supported by the Ramon y Cajal Programme (RYC-2015-18136). N. Díaz-Rodríguez is currently supported by the Spanish Government Juan de la Cierva Incorporación contract (IJC2019-039152-I).

\bibliographystyle{plain}
\bibliography{refs}

\newpage

\section{Appendix}
In this appendix we extend the results obtained with MonuMAI dataset to a second dataset, PASCAL-Part, to further validate our results. We detail such datasets and the results obtained in the following sections.

\subsection{Additional results: PASCAL-Part Dataset} %

In order to validate results with more than one part-based dataset, we expanded experiments to use an adapted version of PASCAL-Part \cite{chen2014detect} which provides  two level of annotations: element annotations for the detection level (\textit{object-parts}), and the macro (\textit{whole}) level with image level labels. %

\textbf{PASCAL-Part Dataset}

PASCAL VOC 2010 dataset is a popular dataset for the task of object detection. It is organized into 20 object classes \cite{everingham2010pascal}. PASCAL-Part dataset extends PASCAL VOC-2010 dataset  with  additional annotations by providing segmentation masks for each object part \cite{chen2014detect}. 

In this work, we use a curated version of the PASCAL-Part provided by \cite{donadello2016integration}\footnote{Available online at \href{https://github.com/ivanDonadello/semantic-PASCAL-Part/}{github.com/ivanDonadello/semantic-PASCAL-Part/}.}.  
The idea is to reduce the number of elements of the original PASCAL-Part by collapsing categories together such as "upper arm" and "lower arm" inside a single category "arm". 

We created a second curated version of the dataset, as in the PASCAL-Part there can be several "macro" objects labelled within a single image, whereas we want to consider, to ease evaluation purposes, only images with one image-level label\footnote{Therefore, we discarded all images where there was more than one macro object.}. The total of 1448 remaining images include 20 macro categories and 44 different parts, whose distribution and some samples are shown in previous sections of the Appendix.

PASCAL-Part dataset classes and parts are in the following list. The first element represents each class, and it is followed by its corresponding part classes\footnote{In OWL language, the latter would be placed in their \textit{hasPart} object property range}:

\begin{itemize}
\item  Bird: Torso, Tail, Neck, Eye, Leg, Beak, Animal Wing, Head
\item Aeroplane: Stern, Engine, Wheel, Artifact Wing, Body
\item  Cat: Torso, Tail, Neck, Eye, Leg, Ear, Head
\item  Dog: Torso, Muzzle, Nose, Tail, Neck, Eye, Leg, Ear, Head
\item  Sheep: Torso, Tail, Muzzle, Neck, Eye, Horn, Leg, Ear, Head
\item  Train: Locomotive, Coach, Headlight
\item  Bicycle: Chain Wheel, Saddle, Wheel, Handlebar
\item  Horse: Hoof, Torso, Muzzle, Tail, Neck, Eye, Leg, Ear, Head
\item  Bottle: Cap, Body
\item  Person: Ebrow, Foot, Arm, Torso, Nose, Hair, Hand, Neck, Eye, Leg, Ear, Head, Mouth
\item  Car: License plate, Door, Wheel, Headlight, Bodywork, Mirror, Window
\item  DiningTable: DiningTable 
\item  Pottedplant: Pot, Plant
\item  Motorbike: Wheel, Headlight, Saddle, Handlebar
\item  Sofa: Sofa
\item  Boat: Boat
\item  Cow: Torso, Muzzle, Tail, Horn, Eye, Neck, Leg, Ear, Head
\item  Chair: Chair
\item  Bus: License plate, Door, Wheel, Headlight, Bodywork, Mirror, Window
\item  TvMonitor: Screen, TvMonitor 
\end{itemize}

\begin{figure}[htbp!]
    \centering
    \includegraphics[width=17cm]{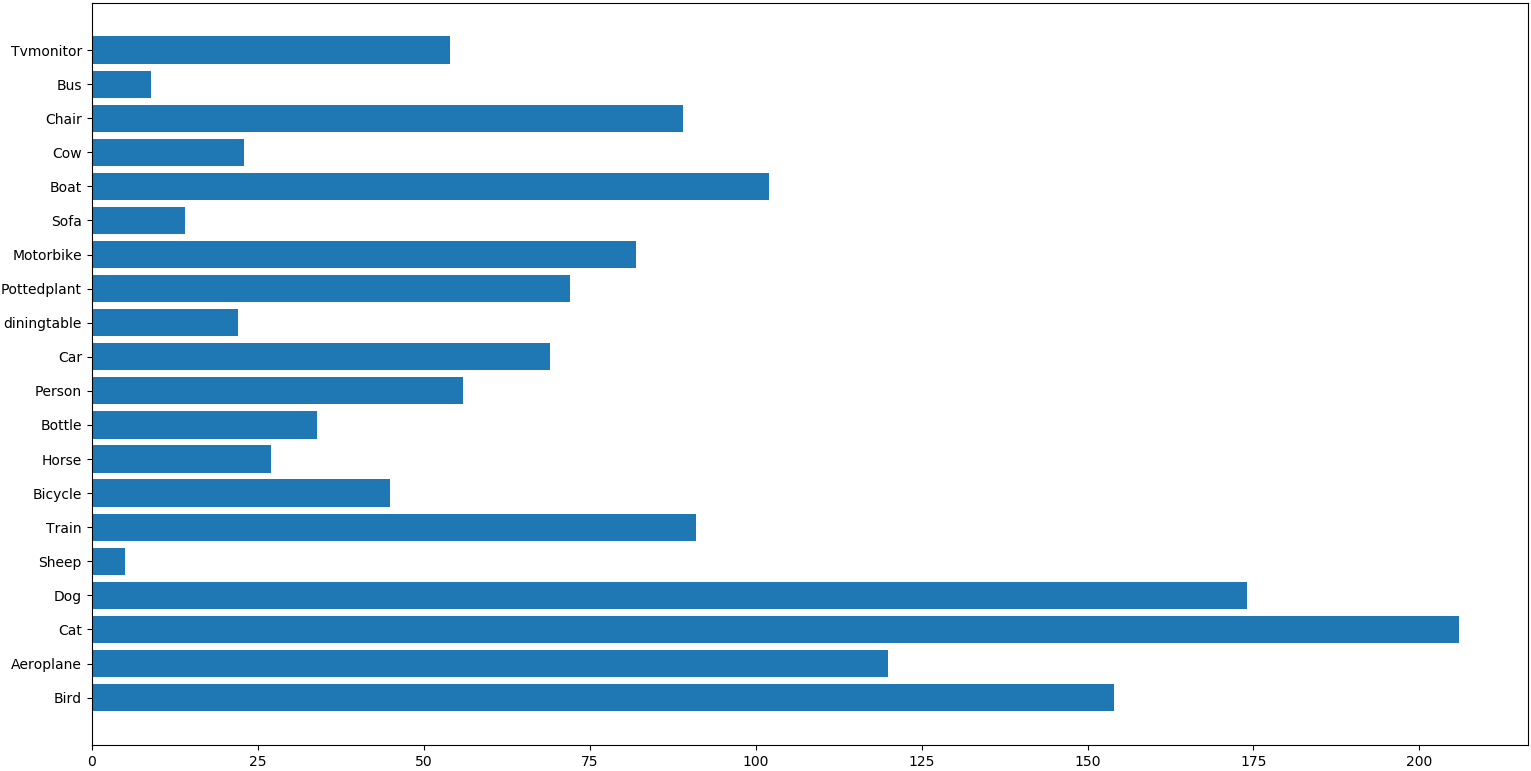}
    \caption{Distribution of classes in PASCAL-Part dataset.}
    \label{fig:label_pascal}
\end{figure}

\begin{figure}[htbp!]
    \centering
    \includegraphics[width=17cm]{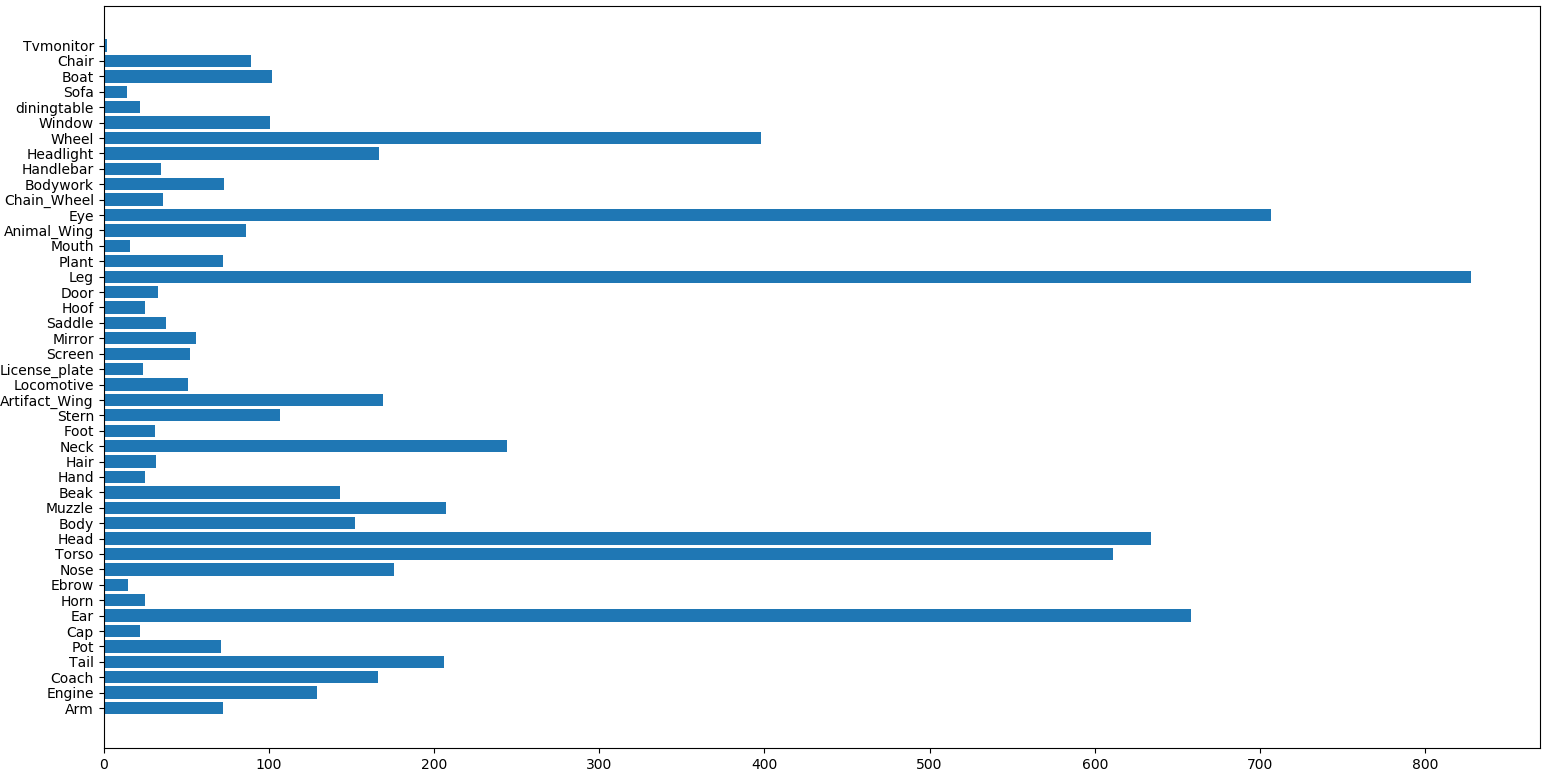}
    \caption{Distribution of object-parts in PASCAL-Part dataset.}
        \label{fig:part_pascal}
\end{figure}

\begin{figure}[htbp!]
\centering
\begin{minipage}{.4\textwidth}
    \centering
    \includegraphics[width=\linewidth]{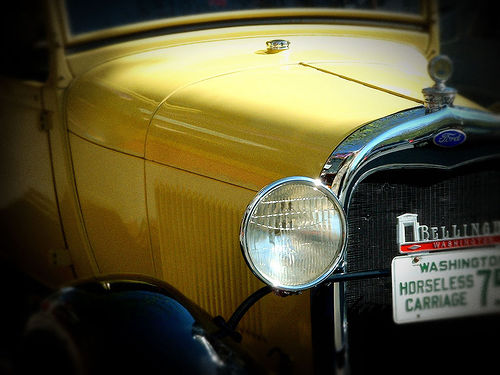}
\end{minipage}%
\begin{minipage}{.4\textwidth}
    \centering
    \includegraphics[width=\linewidth]{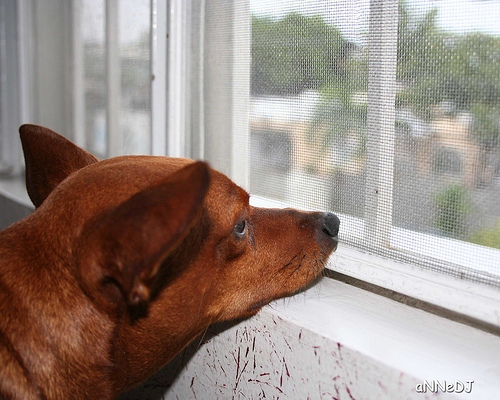}
\end{minipage}
\begin{minipage}{.4\textwidth}
    \centering
    \includegraphics[width=\linewidth]{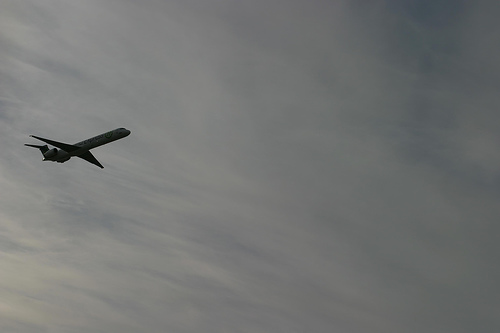}
\end{minipage}%
\begin{minipage}{.4\textwidth}
    \centering
    \includegraphics[width=\linewidth]{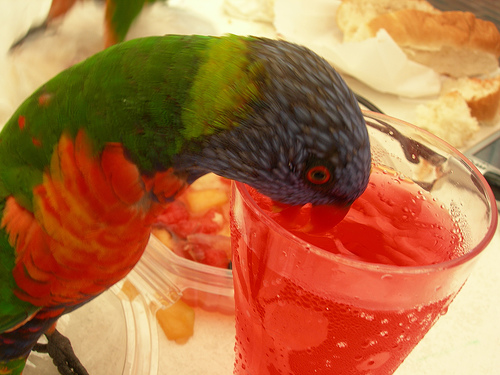}
\end{minipage}
\begin{minipage}{.4\textwidth}
    \centering
    \includegraphics[width=\linewidth]{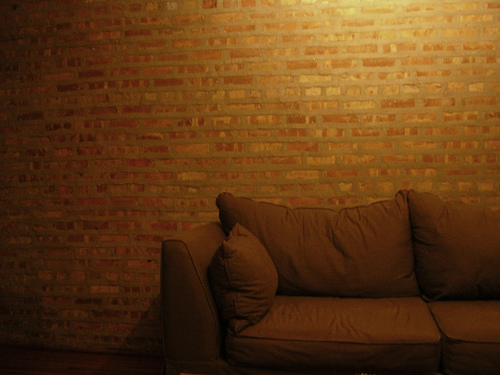}
\end{minipage}
\begin{minipage}{.4\textwidth}
    \centering
    \includegraphics[width=\linewidth]{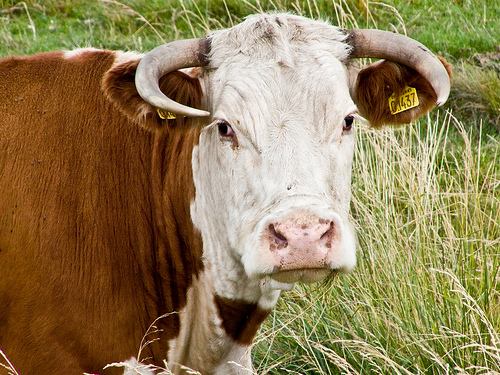}
\end{minipage}
\caption{Extract from PASCAL-Part dataset (I).}
\label{fig:extractPascal1}
\end{figure}

\begin{figure}[htbp!]
\centering
\begin{minipage}{.4\textwidth}
    \centering
    \includegraphics[width=\linewidth]{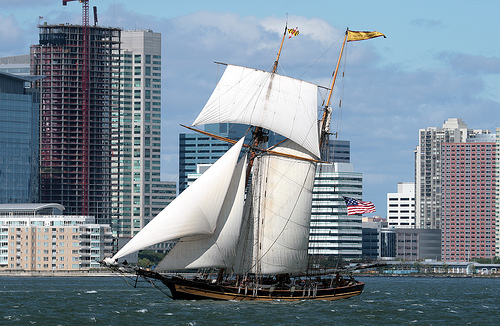}
\end{minipage}
\begin{minipage}{.4\textwidth}
    \centering
    \includegraphics[width=\linewidth]{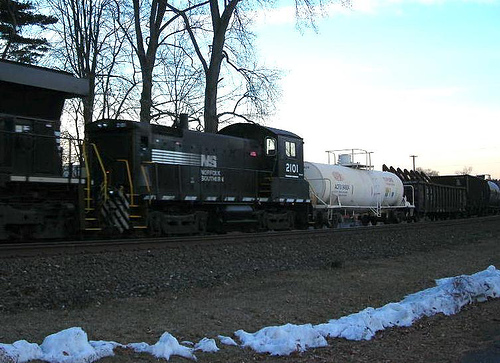}
\end{minipage}
\begin{minipage}{.4\textwidth}
    \centering
    \includegraphics[width=\linewidth]{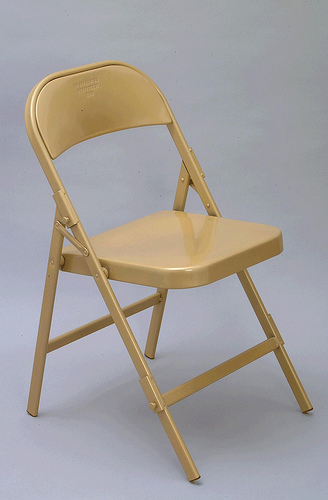}
\end{minipage}
\begin{minipage}{.4\textwidth}
    \centering
    \includegraphics[width=\linewidth]{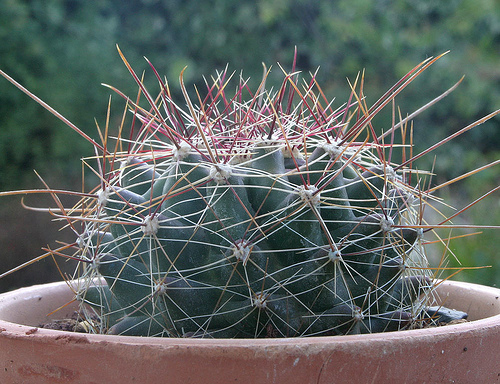}
\end{minipage}
\begin{minipage}{.4\textwidth}
    \centering
    \includegraphics[width=\linewidth]{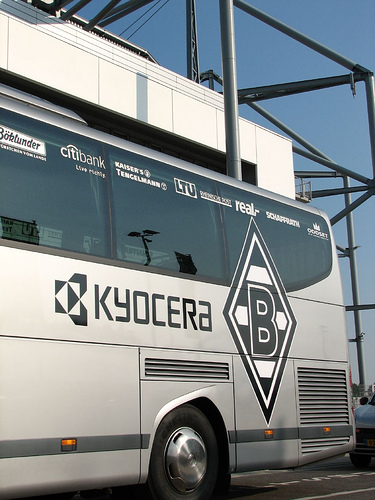}
\end{minipage}
\begin{minipage}{.4\textwidth}
    \centering
    \includegraphics[width=\linewidth]{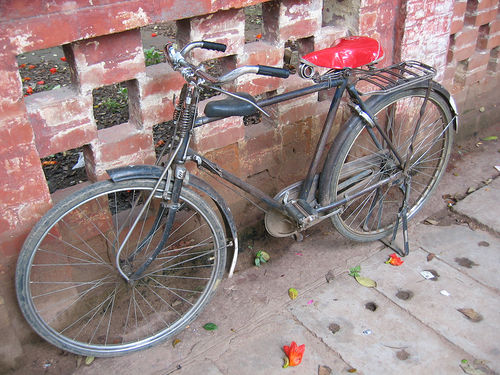}
\end{minipage}
\caption{Extract from PASCAL-Part dataset (II).}
\label{fig:extractPascal2}
\end{figure}

Labels for PASCAL-Part dataset used are shown in Figs. \ref{fig:label_pascal} and \ref{fig:part_pascal}, and image samples are in Figs. \ref{fig:extractPascal1} - %
\ref{fig:extractPascal2}.

Since explicitly using the ontology built for MonuMAI, in the case of object classification yielded no significant advantage,%
we did not pursue this direction further for PASCAL-Part dataset. A thorough work to convert PASCAL-Part into an ontology could be done, but the variety of elements inside it could make it difficult to 1) group them in meaningful categories, 2) extend the additional data and object properties of such richer ontology to a KG that can be compared with an attribution graph. Since such extension to an ontology can complicate the ontology - misattribution matching process, we leave such extension to future work. %

\begin{figure}[htbp!]
\centering
\begin{minipage}{.4\textwidth}
    \centering
    \includegraphics[width=\linewidth]{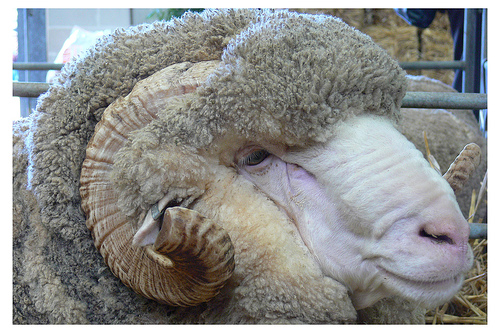}
\end{minipage}
\begin{minipage}{.4\textwidth}
    \centering
    \includegraphics[width=\linewidth]{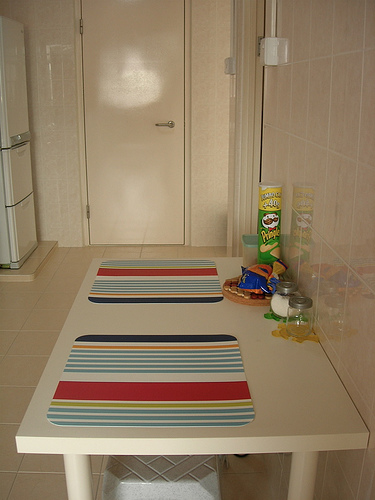}
\end{minipage}
\begin{minipage}{.4\textwidth}
    \centering
    \includegraphics[width=\linewidth]{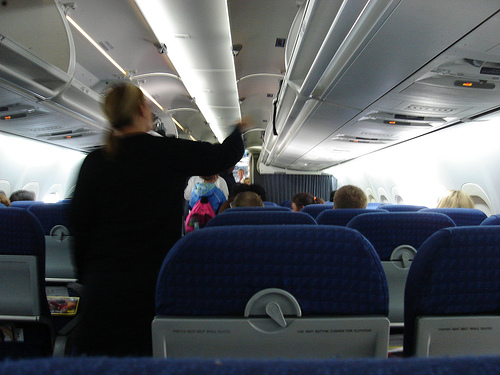}
\end{minipage}
\begin{minipage}{.4\textwidth}
    \centering
    \includegraphics[width=\linewidth]{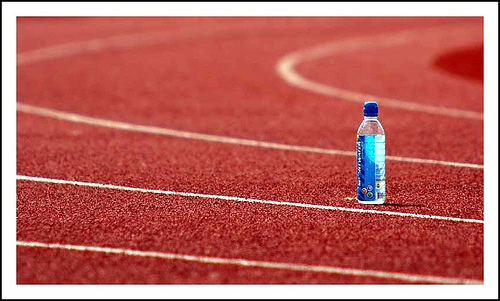}
\end{minipage}
\begin{minipage}{.4\textwidth}
    \centering
    \includegraphics[width=\linewidth]{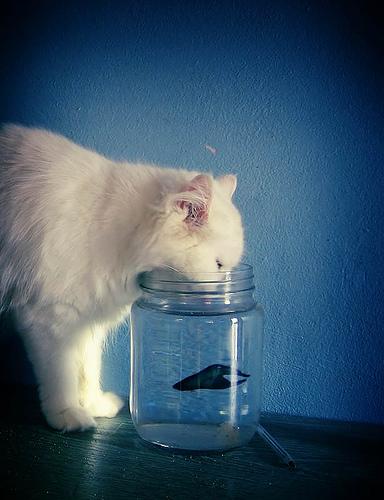}
\end{minipage}
\caption{Extract from PASCAL-Part dataset (III).}
\label{fig:extractPascal3}
\end{figure}

\begin{figure}[htbp!]
\centering
\begin{minipage}{.4\textwidth}
    \centering
    \includegraphics[width=\linewidth]{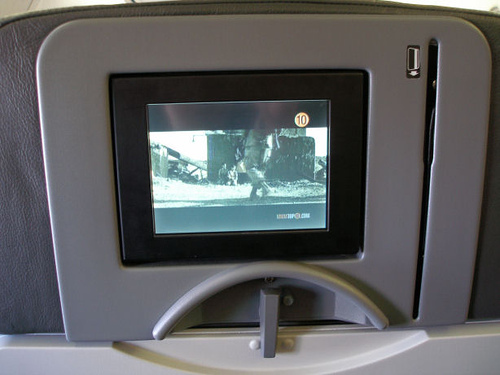}
\end{minipage}
\begin{minipage}{.4\textwidth}
    \centering
    \includegraphics[width=\linewidth]{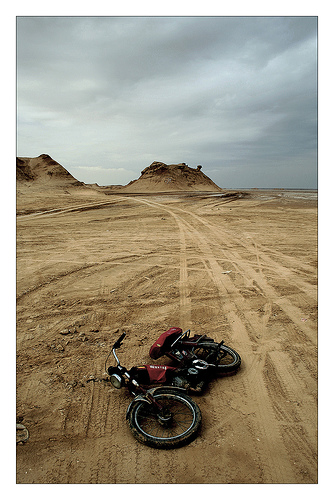}
\end{minipage}
\begin{minipage}{.4\textwidth}
    \centering
    \includegraphics[width=\linewidth]{data/003850.jpg}
\end{minipage}
\begin{minipage}{.4\textwidth}
    \centering
    \includegraphics[width=\linewidth]{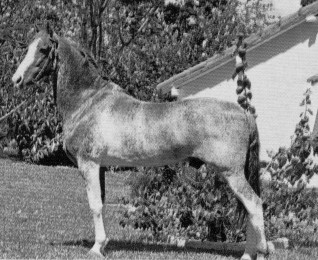}
\end{minipage}
\caption{Extract from PASCAL-Part dataset (IV).}
\label{fig:extractPascal4}
\end{figure}

\textbf{PASCAL-Part's Knowledge Graph}
\label{sec:pascalPart_KG}

  In analogy to MonuMAI previous application, where architectural elements play the role of object parts and macro labels correspond to the architectural styles, we also used the KG provided by the PASCAL-Part dataset \cite{donadello2016integration}\footnote{Curated PASCAL-Part Dataset and KG available \href{https://github.com/ivanDonadello/semantic-PASCAL-Part/}{github.com/ivanDonadello/semantic-PASCAL-Part/}. We do not  provide a visualization of this KG as it would be unreadable.}.

\subsection{Results for PASCAL-Part Dataset}
Results for \textit{PASCAL-Part} dataset are compiled in Table \ref{tab:shap_backprop_results_PASCAL-Part}. Applying SHAP-Backprop has almost no effect on the accuracy and interpretability, but it had detrimental effect on the detector mAP. This could be explained due to this particular KG being very sparsely populated, and the fact that object parts have large overlap in the object labels they theoretically contribute to. For instance, consider the following concrete example: %
We have a person in an image, for which we detect the legs, and let us assume the background is such that sheep legs are detected. According to our expert KG data, detecting legs makes sense toward predicting a person, and thus the sheep legs detection inside the background would not be discouraged. %
Future work should consider the distinction among syntactically equal (e.g. in image captioning tasks, the word \textit{leg}) but semantically different parts of objects (an animal vs a human leg). In other words, the \textit{isPartOf} relationship could be further specialized in our KG to a) have as range the class Leg, with subclasses \textit{AnimalLeg} and \textit{HumanLeg} (instead of just Leg as PASCAL-Part dataset has it now), or b) having \textit{isPartOfAnimal, isPartOfHuman} as extra specialized object properties in our ontology (right now PASCAL-Part and MonuMAI only have one kind of object property, \textit{hasPart}).

The overall lower score in mAP we obtain for \textit{PASCAL-Part} stems from the fact that this dataset makes it harder for smaller objects to be detected, and the Faster R-CNN model we used was not fine-tuned to be fairly comparable in both settings. %

\begin{table}[htbp!]
    \centering
    \begin{tabular}{c|c|c|c}
    & \textbf{mAP} & \textbf{Accuracy} & \textbf{SHAP GED} \\
    & (detection) & (classification) & (interpretability) \\
    \hline
    EXPLANet using Faster R-CNN backbone detector &  $36.5$ & $82.4$ & \textbf{0.45}\\ 
    \hline
    EXPLANet using RetinaNet backbone detector  &  \textbf{39.3} & $80.3$ & $0.59$\\
     \hline
    ResNet-101 baseline classifier & N/A & \textbf{87.2} & N/A\\ 
    \end{tabular}
    \caption{Results of two backbone variants of EXPLANet (using Faster R-CNN and RetinaNet) on PASCAL-Part dataset.}
    \label{tab:result2}
\end{table}

When considering the different weighting schemes of SHAP-Backprop, for the PASCAL-Part, %
the vanilla ResNet classifier baseline performs better %
than that one for EXPLANet, which means that the object-part-based classifier EXPLANet underperforms in this case. It can be explained by several factors, but the predominant one is probably that images contain valuable information that the part-label model does not. It becomes quite clear when studying the PASCAL-Part KG, as we do in Section \ref{sec:pascalPart_KG}, since several labels are made of the same part names, but represent distinct things, i.e., parts from different object provenance (e.g. \textit{leg} of a person and leg of an animal; both \textit{car} and \textit{bus} have the same object-parts). The part-based model has thus trouble differentiating such categories whereas a purely image-based model (and not attribute based) would have no issue with those.

\begin{table}[htbp!]
    \centering
    \begin{tabular}{c|c|c|c}
        & \textbf{mAP}  & \textbf{Accuracy} & \textbf{SHAP GED}  \\
     & (detection) & (classification) & (interpretability) \\
        \hline
        Standard procedure (baseline, no SHAP-Backprop) &  $\textbf{36.5}$ & $\textit{82.4}$ & $\textit{0.45}$\\
        \hline
        Linear BBox-level weighting & $33.0$ & $79.7$ & $0.65$\\
        \hline
        Exponential BBox-level weighting & $32.0$ & $81.0$ & $0.55$  \\
        \hline
        Linear instance-level weighting & $33.4$ & $80.3$ & $\textbf{0.42}$\\
        \hline
        Exponential instance-level weighting & $\textit{34}$ & $\textbf{82.7}$ & $0.56$  \\
    \end{tabular}
    \caption{Impact of the new SHAP-Backprop training procedure on the mAP (detection), accuracy (classification) and SHAP GED (interpretability). Results on PASCAL-Part dataset with EXPLANet architecture. GED is computed between the SAG and its projection in the expert KG. (Standard procedure: sequential typical pipeline of 1) train detector, 2) train classifier without SHAP-Backprop)}
    \label{tab:shap_backprop_results_PASCAL-Part}
\end{table}
While the linear weighting appears to have a more positive effect on improving explainability of the model, it may not be significant, given that it does not always improve interpretability when applied on the more specific  (but with lack of signal)    PASCAL-Part dataset.

The discordance in performance (for mAP and Accuracy in the detector task) in Table \ref{tab:shap_backprop_results_monumai} and \ref{tab:shap_backprop_results_PASCAL-Part} for RetinaNet being superior than Faster R-CNN only in MonuMAI but not for PASCAL-Part can be explained due to PASCAL-Part dataset labelling procedure (joining elements not unifiable with the same identifier). For instance: %
the same label is used for object parts of different semantic and visual meaning: \textit{sheep}'s legs and \textit{cow}' legs, or car's wheel and bike's wheel. 
Therefore, it is worth highlighting the differences in the labelling process of both datasets, as the classification based in parts with the same name but different semantics and visual appearance in PASCAL-Part is not designed for a neural network that only takes attributes as input. %
Thus, as PASCAL-Part lacks highly discriminative features, accuracy and SHAP GED are not obviously nor directly connected, specially in RetinaNet where, due to its inherent architecture aggregation function, all probabilities are used (i.e., summed) to perform a prediction, %
not just the %
highest one. %

As micro and macro level labels are not appropriate (as designed in MonuNet dataset), the interpretability metric fails to reflect reality%
, independently of the quality of the detector.

The take away message from the ablation study assessing the impact of the object detector on EXPLANet (Table \ref{tab:result2}) is that even if X-NeSyL methodology showed slightly differently results in datasets designed with different purpose\footnote{Independently assessed outside EXPLANet, RetinaNet is slightly superior to Faster R-CNN.}, having worse default results when using EXPLANet with RetinaNet instead of with Faster R-CNN could be due to 1) hyperparameter choice, since Faster R-CNN uses pretraining on MS-COCO while RetinaNet uses pretraining on ImageNet, and 2) both coarse grained MonuMAI dataset and fine-grained PASCAL-Part are of different nature in terms of the overlap among \textit{part} classes.
Lacking discriminative dataset labels %
in PASCAL-Part results in insufficient signal for EXPLANet to leverage. Thus, dataset label noise is not appropriate to capture the compositional part-based hierarchy in the data nor to 100\% assess SHAP-Backprop.   %

\end{document}